\documentclass[journal,12pt,onecolumn,draftclsnofoot,]{IEEEtran}
\usepackage{amsmath,amsfonts}
\usepackage{algorithmic}
\usepackage{algorithm}
\usepackage{array}
\usepackage[caption=false,font=tiny,labelfont=sf,textfont=sf]{subfig}
\usepackage{textcomp}
\usepackage{stfloats}
\usepackage{url}
\usepackage{verbatim}
\usepackage{graphicx}
\usepackage{bm}
\usepackage[backend=biber, style=ieee, maxnames=6, minnames=1]{biblatex}
\AtNextBibliography{\footnotesize}
\usepackage{amssymb}
\usepackage{hyperref}
\hypersetup{hidelinks}
\usepackage{multirow}
\begin{document}

\title{Resource-Constrained Edge AI with Early\\Exit Prediction}

\author{Rongkang Dong, Yuyi Mao, \emph{Member IEEE}, and Jun Zhang, \emph{Fellow IEEE}

\thanks{R. Dong and Y. Mao are with the Department of Electronic and Information Engineering, The Hong Kong Polytechnic University, Hong Kong, China (e-mail: rongkang.dong@connect.polyu.hk, yuyi-eie.mao@polyu.edu.hk). (\emph{Corresponding author: Y. Mao}.)\\\indent
J. Zhang is with the Department of Electronic and Computer Engineering, The Hong Kong University of Science and Technology, Hong Kong, China (e-mail: eejzhang@ust.hk).}}

\maketitle

\begin{abstract}
By leveraging the data sample diversity, the early-exit network recently emerges as a prominent neural network architecture to accelerate the deep learning inference process. However, intermediate classifiers of the early exits introduce additional computation overhead, which is unfavorable for resource-constrained edge artificial intelligence (AI). In this paper, we propose an early exit prediction mechanism to reduce the on-device computation overhead in a device-edge co-inference system supported by early-exit networks. Specifically, we design a low-complexity module, namely the Exit Predictor, to guide some distinctly ``hard" samples to bypass the computation of the early exits. Besides, considering the varying communication bandwidth, we extend the early exit prediction mechanism for latency-aware edge inference, which adapts the prediction thresholds of the Exit Predictor and the confidence thresholds of the early-exit network via a few simple regression models. Extensive experiment results demonstrate the effectiveness of the Exit Predictor in achieving a better tradeoff between accuracy and on-device computation overhead for early-exit networks. Besides, compared with the baseline methods, the proposed method for latency-aware edge inference attains higher inference accuracy under different bandwidth conditions.
\end{abstract}


\section{Introduction}
\label{section introduction}
With the great demand of Internet of Things (IoT) applications to embrace the fifth-generation (5G) era, the number of connected devices (e.g., smartphones, surveillance cameras, and unmanned vehicles) has seen a dramatic increase over the past decade, which results in a vast volume of data being generated at the wireless networks \cite{5GIoTsurvey}. Such massive data along with the recent success of deep learning has hastened numerous intelligent mobile applications such as vehicle detection and virtual reality \cite{DLwithECreview, EdgeIntelligence}. However, in order to advance the state-of-the-art accuracy performance, a deep neural network (DNN) typically has tens or even hundreds of network layers, leading to significant computational cost and memory consumption. As a result, it is hard to deploy the whole DNN on resource-constrained devices. Although the high transmission rate offered by the next generation wireless networks allows mobile devices to offload inference tasks to a powerful cloud server, the long communication latency caused by data transmission in the public network prohibits real-time responsiveness \cite{ECsurvey}.
\begin{figure}[h]
  \centering
  \includegraphics[width=0.6\textwidth]{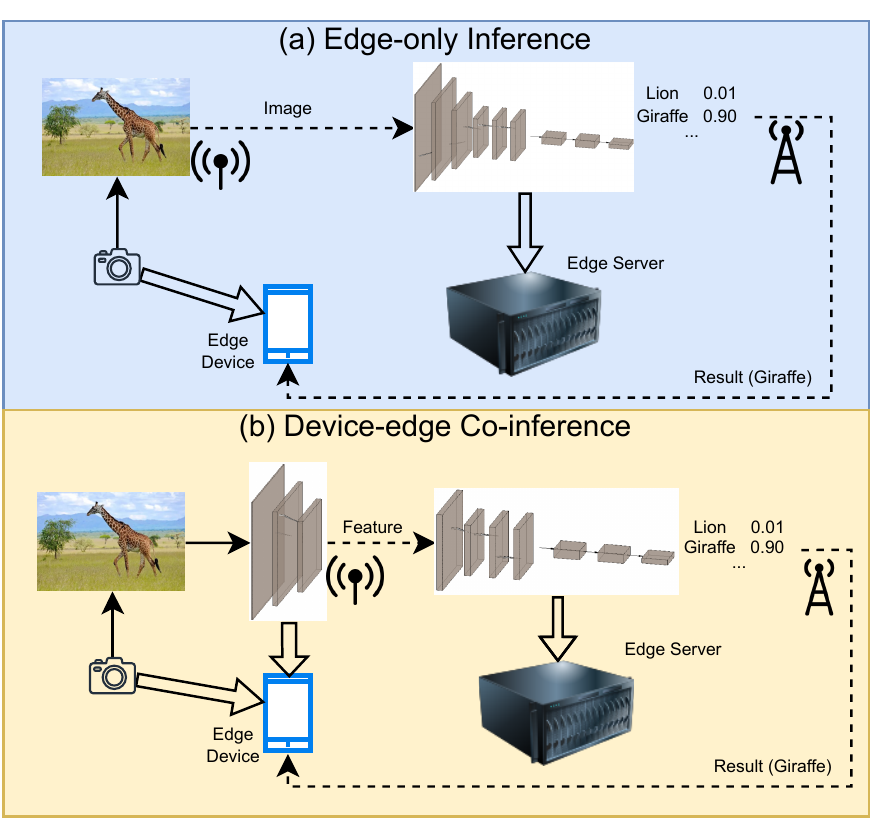}\\
  \centering\caption{Two typical edge inference modes: (a) Edge-only inference and (b) device-edge co-inference.}
  \label{fig: EI modes}
\end{figure}

Edge inference, which deploys mini-servers for computation at the edge of wireless networks (a.k.a. edge servers), is a complementary solution of cloud inference to provide low-latency artificial intelligence (AI) services \cite{ECsurvey, EdgeAI}. Two typical edge inference modes are depicted in Fig. \ref{fig: EI modes}. In particular, \emph{edge-only inference} refers to the scenario that an edge device (e.g., a smartphone) transfers the raw inference input data to an edge server for remote inference and retrieves the inference results thereafter (Fig. \ref{fig: EI modes} (a)). Due to the limited wireless resources and the potentially large size of the inference input, the transmission latency may be too high, especially for applications with tight latency budgets (e.g., autonomous driving \cite{EIonautonomousdriving}). \emph{Device-edge co-inference} is a more agile paradigm that can alleviate this problem. In this edge inference mode, an edge device first extracts a compact intermediate feature by executing the front layers of a backbone DNN, which is then offloaded to the edge server for further processing of the remaining layers \cite{EIlastmile} (Fig. \ref{fig: EI modes} (b)). By making the most use of the computational resources at both the edge device and edge server, device-edge co-inference effectively trims the communication overhead, thus becoming an ideal solution for resource-constrained edge inference \cite{shaotradeoff, neurosurgeon, DDNN, zhangxinjieautoML}.

\begin{figure}[]
  \centering
  \includegraphics[width=0.6\textwidth]{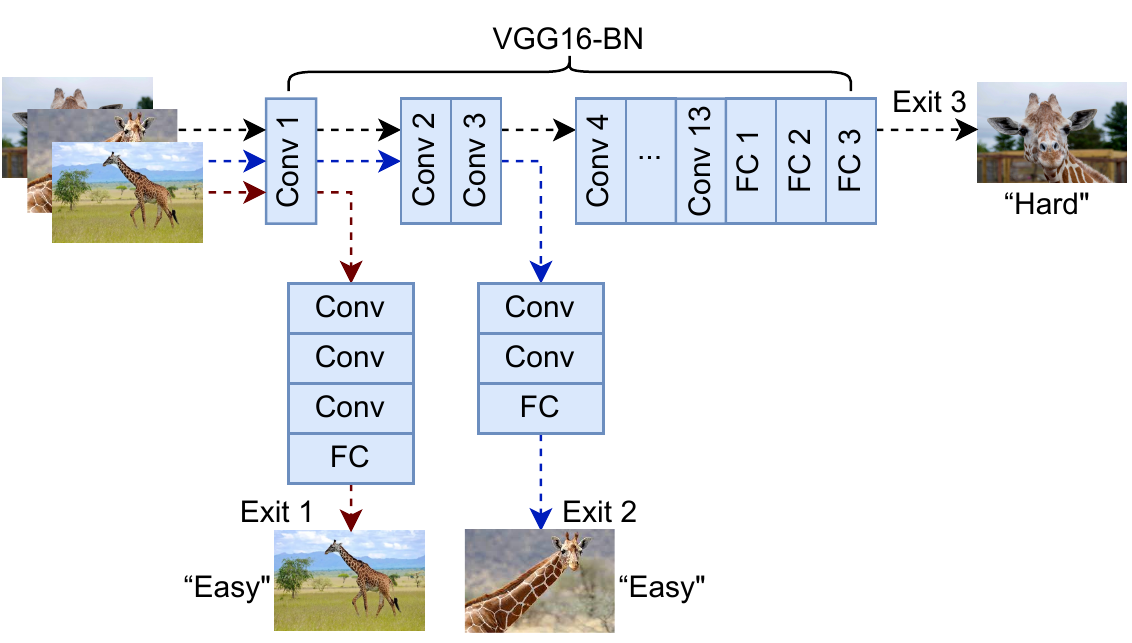}\\
  \caption{An early-exit network with the VGG16-BN \cite{vgg} as backbone. For conciseness, some convolutional layers shown in the figure may also contain a batch normalization (BN) layer, a nonlinear layer, and a pooling layer.}
  \label{fig: VGG16-BN example}
\end{figure}

\begin{figure}[]
  \centering
  \includegraphics[width=0.5\textwidth]{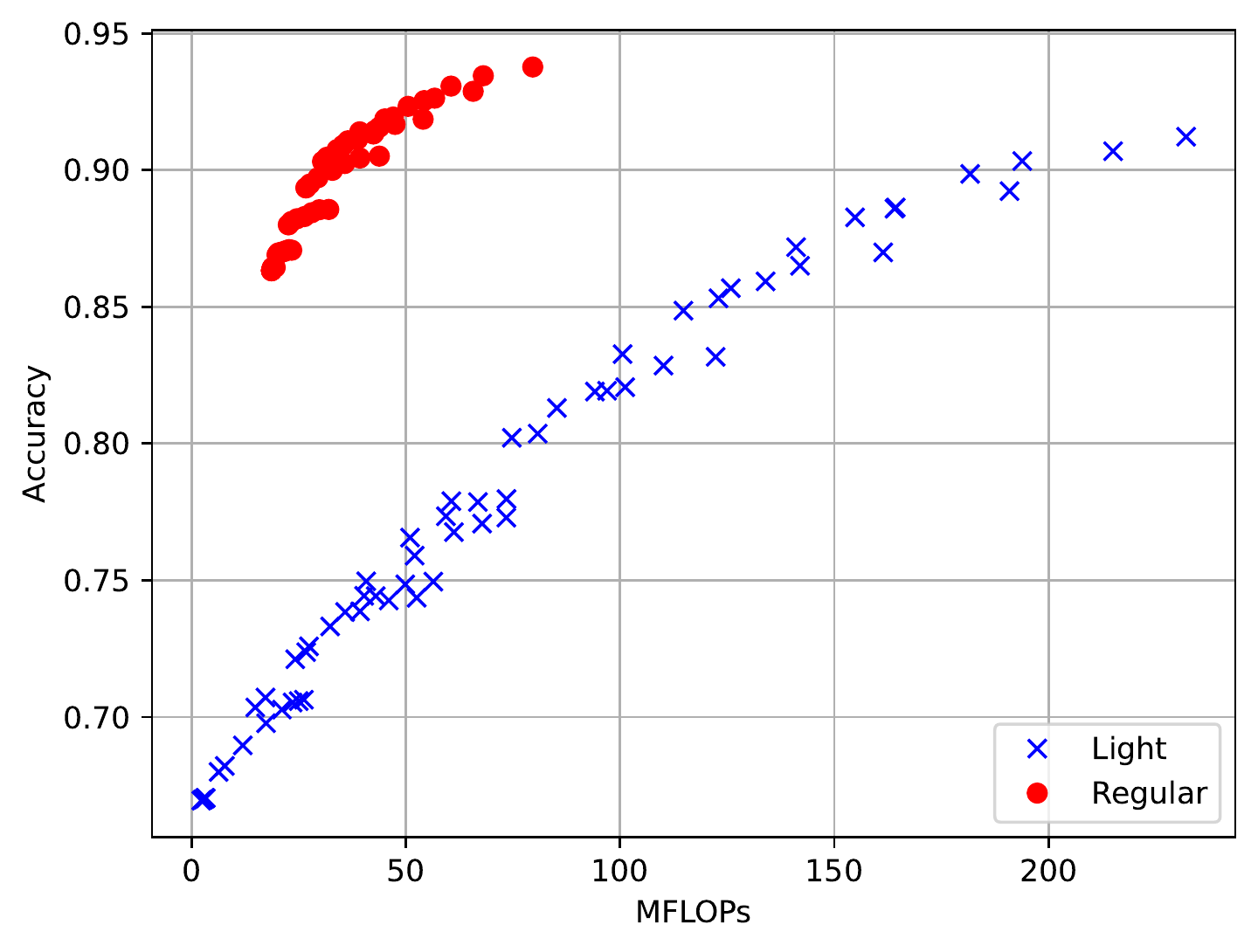}\\
  \caption{The relationship between the inference accuracy and computation complexity (measured by million floating-point operations (MFLOPs)) of the early-exit network shown in Fig. \ref{fig: VGG16-BN example} with regular-size and light intermediate classifiers on the CIFAR10 dataset. The regular-size intermediate classifiers follow the design in Fig. \ref{fig: VGG16-BN example} and the light intermediate classifiers consist of a fully-connected layer after the pooling operation for each exit \cite{lightweightEE}. The inference accuracy and computation complexity tradeoff is adjusted by tuning the confidence thresholds of the early-exit network, as will be elaborated in Section \ref{section system model and design problem}, i.e., $\lambda_{1}, \lambda_{2} \in [0.2, 0.9]$ with a step size of 0.1.}
  \label{fig: light vs regular}
\end{figure}
To enable more efficient device-edge co-inference, the early-exit network, which takes advantages of the \emph{data sample diversity}, has gained increasing interest \cite{branchynet, branchyGNN, DDNN, edgent}. Specifically, an early-exit network can be derived by inserting a few intermediate classifiers along the depth of a backbone network, e.g., VGG16-BN \cite{vgg} as shown in Fig. \ref{fig: VGG16-BN example}. As such, the ``easy" data samples can be inferred by the intermediate classifiers with sufficient confidence, leaving the ``hard" data samples to be processed by all the layers in the main branch. It was shown in \cite{branchynet} that the early exit mechanism can achieve a $5.4\times$/$1.5\times$/$1.9\times$ CPU speedup for the LeNet/AlexNet/ResNet. Early-exit networks can also reduce the communication overhead in device-edge co-inference systems because the ``easy'' samples can be inferred from the early exits deployed on-device without being processed by the server-based network. However, depending on the backbone architecture and the insertion position, an intermediate classifier normally contains at least one convolutional layer and one fully-connected (FC) layer \cite{EEdesign, branchynet, edgent}, which introduces substantial computation overhead. Despite the computation complexity of early-exit networks can be reduced by using light intermediate classifiers (such as those in \cite{lightweightEE}), it comes at a price of severe accuracy degradation especially for data samples with their inference processes being early terminated. To illustrate this issue, we show the relationship between the inference accuracy and computation complexity of a VGG16-BN-based early-exit network in Fig. \ref{fig: light vs regular}. The result shows that the early-exit network with regular-size intermediate classifiers generally requires a lower computation cost for a given accuracy requirement, compared to that with light intermediate classifiers (except with a very low accuracy requirement, i.e., $<$70\%). This is because the light intermediate classifiers fail to terminate the inference process with enough confidence and thus more samples have to be processed by all the backbone layers in order to attain high accuracy.

To reduce the computation overhead brought by the early exits, in this paper, we advocate developing a low-complexity early exit prediction mechanism, named Exit Predictor, for resource-constrained edge AI. Our basic idea is that notable on-device computation can be saved by skipping appropriate early exits for some distinctly ``hard" data samples.

\subsection{Related Works and Motivations}
\label{subsection: Related Works and Motivations}
Collaborative inference, as shown in Fig. \ref{fig: EI modes} (b), is an effective inference model for resource-constrained edge AI \cite{EIlastmile}. In particular, NeuroSurgeon \cite{neurosurgeon} first proposed to partition a DNN model between an edge device and a cloud server, and it attempted to find an optimal model partition point considering both the latency and energy performance. To reduce the storage and accelerate the co-inference process, mixed-precision DNN models are adopted for collaborative inference in \cite{quantizationandpartition}. However, DNN partitioning may result in increased communication overhead because of the in-layer data amplification phenomenon \cite{bottlenet++}, and therefore intermediate feature compression techniques were also exploited to achieve communication-efficient device-edge co-inference \cite{featurecompression, bottlenet++, informationbottleneck}.  

The early-exit network architecture can reduce both the computation and transmission cost of device-edge co-inference \cite{branchynet, surveyonSCandEE}, which avoids all inference data to be processed by every backbone layer. Early-exit networks can also be used in device-edge-cloud collaborative inference systems \cite{DDNN}, where the inference model is divided into three partitions over the distributed computing hierarchy. Besides, Li \emph{et al.} introduced the Edgent framework in \cite{edgent}, which optimizes the latency of device-edge co-inference under static and dynamic bandwidth conditions through adaptive DNN partitioning and right-sizing. Similarly, SPINN \cite{SPINN} optimizes the early-exit policy and model partition point of an early-exit network for cooperative edge inference under dynamic network conditions. Nevertheless, while many prior investigations on early-exit networks focused on improving the model accuracy \cite{dynexit, distillationonEE} and efficiency when being deployed in the edge computing environments \cite{edgent, SPINN}, few studies tackled the additional computational overhead brought by intermediate classifiers \cite{EEdesign}. Although the significant computation overhead of intermediate classifiers has received some most recent attentions, existing solutions either use light intermediate classifiers with severe performance degradation \cite{lightweightEE}, or reuse the predictions of the preceding intermediate classifiers for later ones to gain a better accuracy-time tradeoff \cite{ZeroTimeWaste}. However, the ``hard" samples may still need to be processed by all the intermediate classifiers.
\subsection{Contributions}
\label{subsection: Contributions}
In this paper, we consider a device-edge co-inference system supported by early-exit networks and propose a low-complexity early exit prediction mechanism to improve the on-device computation efficiency. Our major contributions are summarized as follows:
\begin{itemize}
\item We design an innovative early exit prediction mechanism, named Exit Predictor, as an effective means to reduce the on-device computation overhead in device-edge co-inference systems supported by early-exit networks. To make the Exit Predictor with low computation cost, depthwise separable convolution is applied to design its network architecture, meanwhile, the squeeze-and-excitation and channel concatenation operations are adopted to improve the performance. 
\item We extend the Exit Predictor for latency-aware edge inference under different communication bandwidths. To avoid training multiple Exit Predictors for each different bandwidth conditions, we propose to retain just one Exit Predictor, and adapt its hyper-parameters, namely the prediction thresholds, together with the confidence thresholds of the early-exit network via a few simple regression models.
\item We conduct extensive performance evaluation for the proposed methods on image classification tasks. The experiment results show that the Exit Predictor helps reduce the on-device computation overhead remarkably with few accuracy loss or no accuracy loss. For latency-aware edge inference, the proposed method achieves better accuracy compared with other baselines under different bandwidth conditions.
\end{itemize}

\subsection{Organization}
\label{subsection: Organization}
The rest of this paper is organized as follows. In Section \ref{section system model and design problem}, we introduce a device-edge co-inference system supported by early-exit networks and define the design problem for on-device computation overhead reduction. We develop an early exit prediction mechanism in Section \ref{section the proposed early exit prediction mechanism} and extend our investigation to latency-aware edge inference with varying communication bandwidth in Section \ref{section latency-aware early exit prediction}. Experimental results are presented in Section \ref{section experimental results} and conclusions are drawn in Section \ref{section conclusions}.

\section{System Model and Design Problem}
\label{section system model and design problem}
\begin{figure*}[ht]
  \centering
  \includegraphics[width=0.8\textwidth]{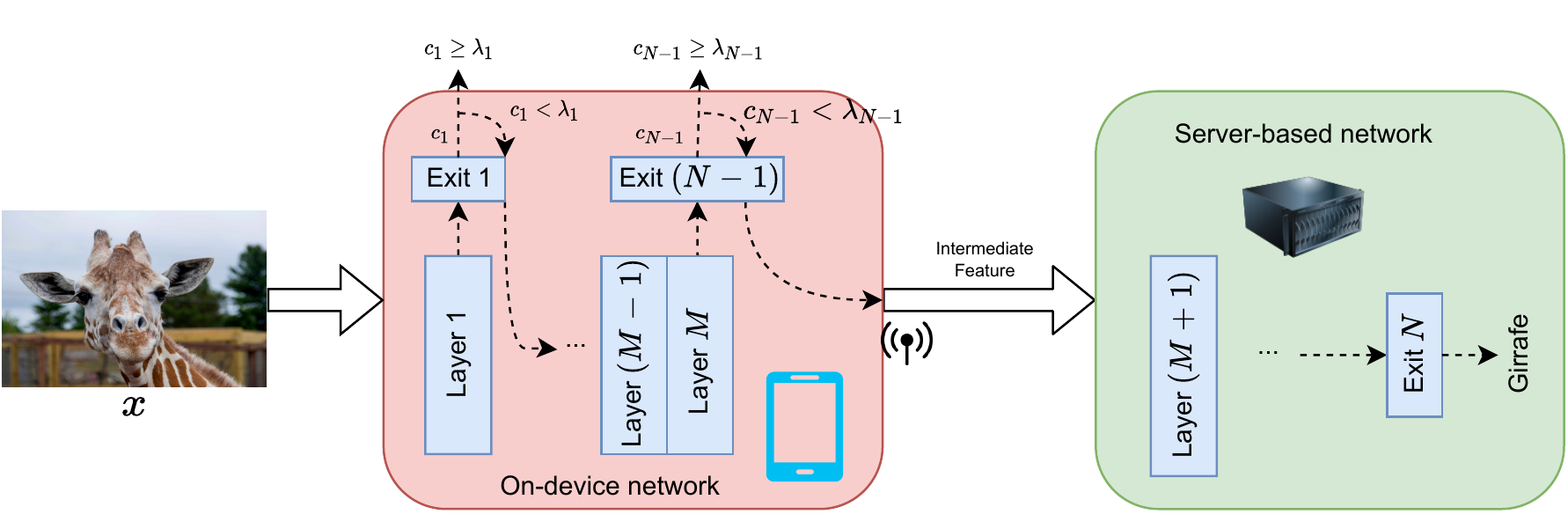}\\
  \caption{A device-edge co-inference system empowered by an early-exit network with $N-1$ early exits.}
  \label{fig: System Model}
\end{figure*}
In this section, we first introduce the device-edge co-inference system supported by early-exit networks. Then, we define the design problem to reduce the on-device computation overhead of the considered edge inference system.

\subsection{Device-edge Co-inference with Early-exit Networks}
\label{subsection: Device-edge Co-inference with Early-exit Networks}
We consider a device-edge co-inference system empowered by an early-exit network as shown in Fig. \ref{fig: System Model}. The early-exit network is derived from a backbone network by inserting $N-1$ early exits (i.e., $N$ exits in total), which is partitioned between an edge device and the edge server as the on-device and server-based networks, respectively. Each early exit contains an intermediate classifier, which may consist of multiple network layers, followed by a fully-connected layer and a soft-max layer \cite{branchynet}. For convenience, the term ``intermediate classifier" is used interchangeably with ``early exit" in the remainder of this paper. We assume that the edge server is well-resourced, i.e., the computation latency at the edge server is negligible compared with those of on-device computation and intermediate feature transmission, and thus all the $N-1$ early exits are deployed on the edge device. Accordingly, the model partition point is chosen as the insertion position of the last early exit.

Each inference data (e.g., an image) is processed sequentially by the $N-1$ early exits until an inference result with sufficient confidence is obtained. It is also possible that none of the early exits is able to obtain a reliable inference result, and in this case, the intermediate feature computed by the on-device network is forwarded to the edge server and the inference result is then determined by the last ($N$-th) exit. We denote the feedforward operation from the input layer to the $n$-th exit as $f_{n}(\cdot)$, and the output of the $n$-th intermediate classifier before the soft-max layer can thus be expressed as follows:
\begin{equation}
    \bm{z}_{n} = f_{n}(\bm{x};\theta_{n}) \in \mathbf{R}^{P}, n=1, 2, ..., N-1,
    \label{equation: early exit feedforward function}
\end{equation}
where $\bm{x}$ is the input data, $\theta_{n}$ encapsulates the network parameters from the input layer to the layer before the soft-max operation in the $n$-th exit, and $P$ represents the number of categories in the classification task (e.g., $P=10$ in the CIFAR10 image classification task). Following most prior studies on early-exit networks, we use the top-1 probability \cite{SPINN} as the confidence score of the intermediate inference results, which can be written as follows:
\begin{equation}
    c_{n} = {\rm max}({\rm softmax}(\bm{z}_{n})), n=1,2, \cdots, N-1,
\end{equation}
where ${\rm softmax}(\bm{z}_{n})\triangleq \{e^{z_{n}^{p}}/\sum_{q=1}^{P}e^{z_{n}^{q}}\}_{p=1}^{P}$ and $z_{n}^{p}$ denotes the $p$-th dimension in vector $\bm{z}_{n}$. If the confidence score $c_{n}$ is no smaller than a pre-determined confidence threshold $\lambda_{n}\in (0, 1)$, the intermediate inference result is deemed with sufficient confidence and the inference process of $\bm{x}$ is terminated at the $n$-th early exit. Otherwise, the feature computed by the backbone layer right before the $n$-th early exit is fed to the next backbone layer for further processing. We note that there is no need to set a confidence threshold for the last exit as the inference process has to be terminated at the edge server if none of the $N-1$ early exits with $c_{n}\geq \lambda_{n}, n=1, 2, ..., N-1$.

\subsection{Design Problem}
\label{subsection: Design Problem}
In the considered device-edge co-inference system, the average on-device computation, which is measured by the number of FLOPs\footnote{The FLOPs of different neural network models can be conveniently obtained in the PyTorch environment by using the ``Flops counter for convolutional networks in pytorch framework" package, which is available online: \href{https://github.com/sovrasov/flops-counter.pytorch}{https://github.com/sovrasov/flops-counter.pytorch}.}, can be expressed as follows:
\begin{equation}
    (O_{l_{1}}+O_{e_{1}})+\mathbf{E}_{p(\bm{x})}\left\{\sum_{n=2}^{N-1}(O_{l_{n}}+O_{e_{n}})\cdot \prod_{i=1}^{n-1}\mathbf{1}_{c_{i}<\lambda_{i}}\right\},
    \label{equation: on-device computation 1}
\end{equation}
where $p(\bm{x})$ denotes the input data distribution and $\mathbf{1}_{E}$ is an indicator function that equals 1 if event $E$ happens and 0 if otherwise. Besides, $O_{e_{n}}$ denotes the number of FLOPs required to compute the $n$-th intermediate classifier and $O_{l_{n}}$ gives the number of FLOPs required to compute the backbone layers between the $(n-1)$-th and $n$-th early exit ($n=0$ represents the input layer). For the example in Fig. \ref{fig: VGG16-BN example}, $O_{l_{2}}$ is the number of FLOPs required to compute the ``Conv 2" and ``Conv 3" layers, and $O_{e_{2}}$ is the number of FLOPs required to compute the two convolutional layers and one fully-connected layer in Exit 2.

It is clear from (\ref{equation: on-device computation 1}) that those samples being terminated by the $n$-th exit need to be processed by all the $n-1$ preceding intermediate classifiers. Therefore, many of the on-device computations spent on the first few early exits are wasted if a large number of input samples are terminated by the deeper exits. In the next section, we develop an early exit prediction mechanism to reduce the on-device computation overhead by wisely skipping the processing of some early exits. 

\section{The Proposed Early Exit Prediction Mechanism}
\label{section the proposed early exit prediction mechanism}
\begin{figure}[t]
  \centering
  \includegraphics[width=0.6\textwidth]{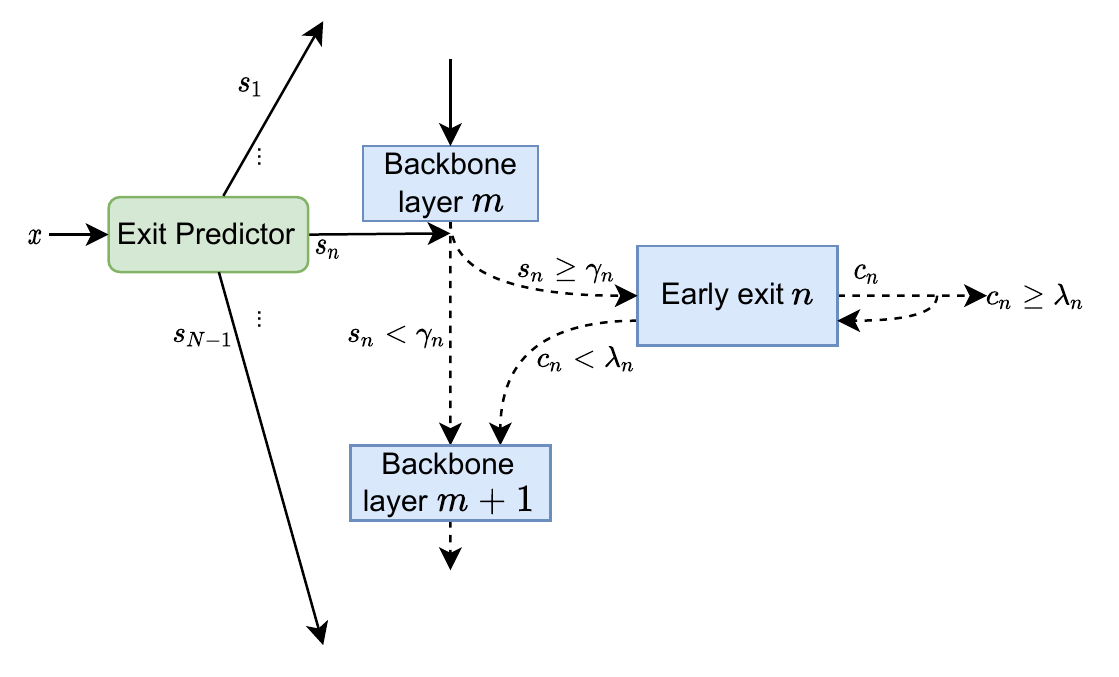}\\
  \caption{Working mechanism of the Exit Predictor.}
  \label{fig: mechanism}
\end{figure}
In this section, we develop an early exit prediction mechanism, named Exit Predictor, to save the on-device computation for resource-constrained edge inference. To justify our motivation, we consider an instance of the early-exit network shown in Fig. \ref{fig: VGG16-BN example} with both of the two confidence thresholds set as 0.9, corresponding to the point in Fig. \ref{fig: light vs regular} with the highest accuracy. For the CIFAR10 test set with 10,000 images, 66.62\%, 19.81\%, and 13.57\% of the input samples are output from the three exits, respectively, leading to 79.64 MFLOPs (including 42.44 MFLOPs on-device). Ideally, if there is an oracle to direct the samples output from the appropriate exit, the required number of FLOPs to process this test set can be reduced to 72.13 MFLOPs (including 34.93 MFLOPs on-device).

The Exit Predictor shall determine whether some early exits can be skipped without being computed for the particular input samples. Our idea is actually similar to those in \cite{IFCNN, dynamicprunelevel}, which aim at selecting the inference network for each input sample via a lightweight neural network, so that some “easy” samples can be processed by the low-complexity models. However, the methods developed in \cite{IFCNN, dynamicprunelevel} are not readily applicable to early-exit networks. On one hand, the lightweight neural network developed in \cite{IFCNN} directly learns the top-1 probabilities for inference model selection. But as this is a complex multivariate regression task, large discrepancies between the predicted and ground-truth values may be incurred. Also, \cite{dynamicprunelevel} proposed to learn the sample complexity for selection between two inference models with mild and aggressive pruning, respectively, which, however, cannot be used for early-exit networks with three or even more early exits.
\subsection{Early Exit Prediction Mechanism}
\label{subsection: Early Exit Prediction Mechanism}
Since the inference results are obtained from the last exit if all the intermediate classifiers fail to derive sufficiently confident results, the proposed Exit Predictor derives $N-1$ prediction scores $s_{n}, n=1, ..., N-1$ as indicators on whether an early exit should be computed or not. Fig. \ref{fig: mechanism} shows the working mechanism of the Exit Predictor, which is also deployed on the device. Specifically, before being fed to the early-exit network, each input sample $\bm{x}$ is first processed by the Exit Predictor to generate the prediction scores $\{s_{n}\}$'s for all the $N-1$ early exits. During the inference process, if $s_{n}$ is no smaller than a predefined prediction threshold $\gamma_{n}$, the intermediate feature computed by the backbone layers is processed by the intermediate classifier in the $n$-th early exit. Otherwise, the $n$-th early exit is skipped and the intermediate feature is passed to the next backbone layer. Nonetheless, for a sample computed by the $n$-th early exit, whether the inference process is actually terminated is still governed by the relationship between the confidence score $c_{n}$ and the confidence threshold $\lambda_{n}$, as elaborated in Section \ref{subsection: Device-edge Co-inference with Early-exit Networks}. The inference procedures of early-exit networks aided by the Exit Predictor are summarized in Algorithm \ref{algorithm: early exit prediction procedure}.
\begin{algorithm}[htb]
\small
\caption{Inference with early-exit networks aided by the Exit Predictor.}
\label{algorithm: early exit prediction procedure}
\textbf{Input:} $\bm{x}$, $\lambda_{n}$, $\gamma_{n}, n=1,\cdots,N-1.$
\begin{algorithmic}[1]
\STATE Compute the predictor scores $\{s_{n}\}$'s for input $\bm{x}$ using the Exit Predictor.
\STATE Compute the on-device backbone layers for input $\bm{x}$ before the first early exit.
\FOR{$n=1$ to $N-1$}
\IF{$s_{n} \geq \gamma_{n}$}
\STATE Compute the $n$-th early exit and obtain the confidence $c_{n}$.
\IF{$c_{n} \geq \lambda_{n}$}
\STATE Return the inference result obtained from the $n$-th early exit and terminate the inference process.
\ELSE 
\STATE Proceed to the next backbone layer.
\ENDIF
\ELSE 
\STATE Bypass the $n$-th early exit and proceed to the next backbone layer.
\ENDIF
\ENDFOR
\STATE Transmit the intermediate feature to the edge server for the processing of the server-based network.
\STATE Return the inference result obtained from the last exit.
\end{algorithmic}
\end{algorithm}

As a result, the average on-device computation with the proposed early exit prediction mechanism can be written in the following expression:
\begin{equation}
    \left( O_{EP}+O_{l_{1}} \right)+\mathbf{E}_{p(\bm{x})} \bigg\{ O_{e_{1}}\cdot \mathbf{1}_{s_{1}\geq\gamma_{1}} +\sum_{n=2}^{N-1}(O_{l_{n}}+O_{e_{n}}\cdot \mathbf{1}_{s_{n}\geq\gamma_{n}})\cdot \prod_{i=1}^{n-1}\left(1-\mathbf{1}_{s_{i}\geq\gamma_{i}}\cdot \mathbf{1}_{c_{i}\geq\lambda_{i}}\right) \bigg\},
\label{equation: on-device computation 2}
\end{equation}
where $O_{EP}$ denotes the number of FLOPs required to compute the Exit Predictor. Since the processing of the Exit Predictor is compulsory for every sample, the Exit Predictor should have a low computational complexity in order not to compromise its advantages of skipping the early exits.

\subsection{Network Architecture}
\label{subsection: Network Architecture}
To implement the Exit Predictor with a lightweight neural network, we formulate the task of generating the $N-1$ prediction scores $\{s_{n}\}$'s as $N-1$ binary classification tasks, where $s_{n}$ can be interpreted as the likelihood that the confidence score $c_{n}$ is no smaller than the pre-defined confidence threshold $\lambda_{n}$.

\begin{figure}[t]
  \centering
  \includegraphics[width=0.4\textwidth]{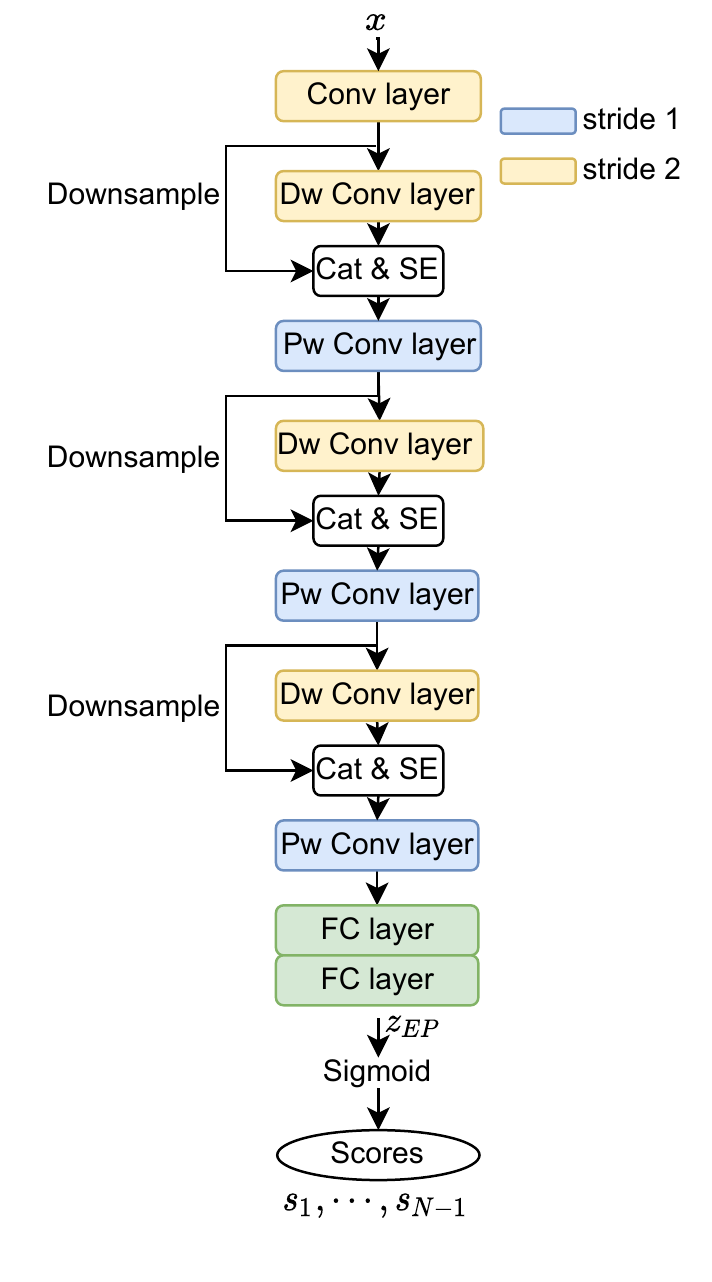}\\
  \caption{Network architecture of the Exit Predictor (``Cat." stands for channel concatenation and ``SE" refers to the squeeze-and-excitation operation. ``Dw" and ``Pw" are the abbreviations of ``depth-wise" and ``point-wise" respectively. The blue blocks are layers with stride 1 and the yellow blocks are layers with stride 2. The green blocks are the fully-connected layers).}
  \label{fig: Exit Predictor}
\end{figure}

Fig. \ref{fig: Exit Predictor} shows the proposed network architecture of the Exit Predictor, and its computational cost and storage requirement mainly come from the convolutional layers and FC layers. In order to reduce the size of convolutional layers, we replace the traditional convolution with the depth-wise separable convolution that consists of sequential depth-wise convolution and point-wise convolution \cite{mobilenet1, mobilenet2, mobilenet3}, except for the first convolutional layer. In our experiment results as shown in Section \ref{section experimental results}, the strides of the first convolutional layer and the depth-wise convolutional layers are chosen as 2, which reduce the height and width of input features by half. In this way, the computational complexity of the Exit Predictor can be significantly reduced. To boost the performance of the Exit Predictor, we concatenate the input and output feature channels \cite{channelconcatenation} of the depth-wise convolutional layers before the squeeze-and-excitation operation \cite{squeezeandexcitation}, and max pooling is adopted as the downsample operation for channel concatenation. The number of channels output from the first convolutional layer is 16 and those output from the three point-wise convolutional layers are 32, 64, and 128, respectively. The last FC layer is cascaded with a Sigmoid function to convert each output dimension to a prediction score from 0 to 1.

\subsection{Training Procedures}
\label{subsection: Training Procedures}
The Exit Predictor imitates the behavior of a pretrained early-exit network. The confidence thresholds for the $N-1$ early exits also need to be determined before training the Exit Predictor. In particular, the early-exit network is trained by optimizing the following loss function on a training dataset $\{\bm{x}_{b}, \bm{y}_{b}\}_{b=1}^{B}$: 
\begin{equation}
    L_{EE}\triangleq\frac{1}{B}\sum_{b=1}^{B}\sum_{n=1}^{N}w_{n}\cdot l(\bm{\hat{y}}_{n,b}, \bm{y}_{b}; \theta_{EE}),
    \label{equation: ee network loss function}
\end{equation}
where $\theta_{EE}$ denotes the network parameters, $w_{n}$ is the weight of the $n$-th exit, $\bm{\hat{y}}_{n,b}\triangleq {\rm softmax}(\bm{z}_{n,b})$, and $\bm{z}_{n,b}$ is the output of the $n$-th exit for the $b$-th training sample before the soft-max layer as given by (\ref{equation: early exit feedforward function}). Besides, the labels $\{\bm{y}_{b}\}$'s are encoded in one-hot format and $l\left(\bm{\hat{y}},\bm{y}\right)\triangleq -\sum_{p=1}^{P}y_{p}\cdot {\rm ln}\hat{y}_{p}$ denotes the cross entropy function. \\\indent
Once the early-exit network is trained, we can collect the data to train the Exit Predictor. Therefore, the training objective of the Exit Predictor is designed as follows:
\begin{equation}
    L_{EP}\triangleq\frac{1}{B}\sum_{b=1}^{B}\sum_{n=1}^{N-1} l_{BCE}(s_{n,b}, u(c_{n,b}-\lambda_{n}); \theta_{EP}),
    \label{equation: EP loss function}
\end{equation}
where $l_{BCE}(\hat{y},y)\triangleq -\left(y\cdot{\rm ln}\hat{y}+(1-y)\cdot{\rm ln}(1-\hat{y})\right)$ is the binary cross entropy loss and $\theta_{EP}$ represents the network parameters of the Exit Predictor. The value of $\bm{s}_{n,b}$ is constrained within 0 to 1 by a Sigmoid function, i.e., the last layer of the Exit Predictor, as follows:
\begin{equation}
    s_{n,b}=\frac{1}{1+e^{-z_{EP_{n,b}}}}
    \label{equation: sigmoid}
\end{equation}
with $z_{EP_{n,b}}$ denoting the $n$-th element of $\bm{z}_{EP_{b}}$, and
\begin{equation}
    \bm{z}_{EP_{b}}=f_{EP}(\bm{x}_{b};\theta_{EP}) \in \mathbf{R}^{N-1}
    \label{equation: exit function}
\end{equation}
as the output of the Exit Predictor of the training sample $\bm{x}_{b}$ before the Sigmoid function. We note that in (\ref{equation: EP loss function}), $u(\cdot)$ is the unit step function and thus $u(c_{n,b}-\lambda_{n})$ given below is the target binary label of sample $\bm{x}_{b}$:
\begin{equation}
    u(c_{n,b}-\lambda_{n})=\left \{
    \begin{aligned}
        1, \quad c_{n,b}\geq\lambda_{n} \\
        0, \quad c_{n,b}<\lambda_{n} \\
    \end{aligned}
    \indent, n=1, 2, ..., N-1. \right
    .
    \label{equation: unit step function}
\end{equation}
\indent When training the Exit Predictor, each sample is first processed by the early-exit network and the confidence scores $\{c_{n,b}\}$'s corresponding to the $N-1$ early exits are generated. Then, the pre-defined confidence thresholds are applied to obtain binary labels as shown in (\ref{equation: unit step function}). Hence, the training objective can be optimized via the back-propagation algorithm \cite{backpropagation}.

\section{Latency-aware Early Exit Prediction}
\label{section latency-aware early exit prediction}
Although the Exit Predictor developed in Section \ref{section the proposed early exit prediction mechanism} is able to reduce the amount of on-device computations, it fails to optimize the critical end-to-end inference latency, which consists of not only the on-device computation latency, but also the transmission latency. This section extends the Exit Predictor to scenarios with varying communication bandwidth, striving for the best tradeoff between the accuracy and latency performance.

To reduce the transmission latency, we compress the intermediate feature prior to transmission \cite{shaotradeoff}. Also, when the communication bandwidth is low, it is desirable to keep more samples with their inference processes being terminated by the early exits on the device, which can be achieved by tuning the confidence thresholds \cite{SPINN, edgeml}. Therefore, the device-edge co-inference system seeks to maximize the inference accuracy subject to an average inference latency requirement $\tau$, which can be formulated as follows:
\begin{equation}
\begin{aligned}
     &\max_{\theta_{EP},\bm{\lambda}, \bm{\gamma}} \quad && Accuracy\left(\hat{\theta}_{EE}, \theta_{EP}, \bm{\lambda}, \bm{\gamma}\right) \\ 
     &\ \ \ \textrm{s.t.} \quad && \mathbb{E}_{p\left(\bm{x}\right)}\left[Latency(\bm{x}, \hat{\theta}_{EE}, \theta_{EP}, \bm{\lambda}, \bm{\gamma})\right] \leq \tau, 
\end{aligned}
\label{equation: optimization}
\end{equation}
where $\hat{\theta}_{EE}$ denotes the pretrained parameters of the early-exit network including the feature encoder and decoder, $\theta_{EP}$ denotes the parameters of the Exit Predictor, and $\bm{\lambda}$ and $\bm{\gamma}$ encapsulate the $N-1$ confidence thresholds and prediction thresholds, respectively. The latency for each sample depends on both the device computation speed $r_{\rm{comp}}$ and communication bandwidth $r_{\rm{comm}}$, and can be calculated according to the following expression:
\begin{equation}
    Latency\left(\bm{x}, \hat{\theta}_{EE}, \theta_{EP}, \bm{\lambda}, \bm{\gamma}\right)= \frac{O_{device}(\bm{x}, \hat{\theta}_{EE}, \theta_{EP}, \bm{\lambda}, \bm{\gamma})}{r_{\rm comp}}
     +\frac{D\left(\bm{x},\hat{\theta}_{EE}, \theta_{EP}, \bm{\lambda}, \bm{\gamma}\right)}{r_{\rm{comm}}},
    \label{equation: latency}
\end{equation}
where $O_{device}\left(\cdot\right)$ denotes the on-device computation and $D\left(\cdot\right)$ is the compressed feature size. Specifically, we follow the idea of Bottlenet++ \cite{bottlenet++} by adopting an autoencoder to compress the intermediate feature before transmission, which is decompressed at the edge server before being processed by the server-based network. We also truncate the feature bit-width to attain further communication overhead reduction. The training procedures of the early-exit network with the autoencoder and quantizer are summarized below:
\begin{itemize}
    \item We first pretrain the early-exit network $\theta_{EE}$ in an end-to-end manner as discussed in Section \ref{subsection: Training Procedures}.
    \item An autoencoder is inserted at the selected model partition point, and it is trained with the server-based network while freezing the on-device network. In this way, the autoencoder causes no effect on the early exits. 
    \item We fine-tune the server-based network after adding the quantization module to increase the accuracy of the last exit. Since the quantization model cannot be back-propagated, the parameters of the on-device network and the feature encoder are frozen in this step.
\end{itemize}

When the confidence thresholds are adjusted according to the communication bandwidth, the Exit Predictor needs to be adapted accordingly as it is trained with a given set of $\{\lambda_{n}\}$'s. A straightforward solution is to train an Exit Predictor for each possible communication bandwidth. However, deploying a large number of Exit Predictors shall result in significant memory footprint, which is not desirable for resource-constrained edge AI. As a result, we propose to retain just one Exit Predictor, which imitates the early-exit network with the highest accuracy\footnote{We perform a multi-dimensional grid search on the confidence thresholds to obtain an early-exit network with the highest accuracy on the test set, and train the Exit Predictor using the procedures in Section \ref{subsection: Training Procedures} on the training set.}, and adapts the prediction thresholds and confidence thresholds to the communication bandwidth. To determine the threshold values under different bandwidth conditions, a few simple regression models (each is made up of two FC layers) are trained, which are facilitated by recording the optimal combinations of the threshold values, i.e., $\bm{\lambda}$ and $\bm{\gamma}$, for (\ref{equation: optimization}) with both $\hat{\theta}_{EE}$ and $\theta_{EP}$ fixed on multiple small sets of discrete communication bandwidth values of the test set\footnote{The number of regression models to be trained and the sets of bandwidth values used for training the regression models are hyper-parameters that need to be tuned in practice.}.

\section{Experimental Results}
\label{section experimental results}
In this section, we evaluate the performance of the proposed early exit prediction mechanism through numerical experiments.

\subsection{Experimental Setup}
\label{subsection: Experimental Setup}
We conduct the experiments with three backbone DNN architectures, including the AlexNet\footnote{Consider the kernel size and stride of the original AlexNet \cite{alexnet} may be too large for $32\times32$ images, we use a modified AlexNet architecture with smaller kernel size and stride, according to the implementation in \href{https://github.com/Lornatang/pytorch-alexnet-cifar100}{https://github.com/Lornatang/pytorch-alexnet-cifar100}.}, VGG16-BN \cite{vgg}, and ResNet44 \cite{resnet} on the CIFAR10 and CIFAR100 datasets \cite{CIFAR}. The two datasets contain 50,000 training samples and 10,000 test samples, which are $32\times32$ color images from 10 and 100 categories, respectively. All our experiments are implemented with the PyTorch library \cite{pytorch}. TABLE \ref{table: backbone accuracy and computation} shows the model accuracy and the computation complexity of the three backbone networks on the two datasets.

\begin{table}[h]
\footnotesize
\centering
\caption{Backbone model accuracy and computation complexity.}
\label{table: backbone accuracy and computation}
\begin{tabular}{c|c|c|c}
\hline
Dataset                   & Backbone & Accuracy(\%) & MFLOPs \\ \hline
\multirow{3}{*}{CIFAR10}  & AlexNet  & 85.95   & 62.88  \\ \cline{2-4} 
                          & VGG16-BN & 94.04   & 333.08 \\ \cline{2-4} 
                          & ResNet44 & 93.91   & 98.52  \\ \hline
\multirow{3}{*}{CIFAR100} & AlexNet  & 60.32   & 63.25  \\ \cline{2-4} 
                          & VGG16-BN & 72.70   & 333.45 \\ \cline{2-4} 
                          & ResNet44 & 72.21   & 98.52  \\ \hline
\end{tabular}
\end{table}

Two early exits are inserted into the on-device network, and the weights in the training objective (\ref{equation: ee network loss function}) for the three exits are set as 0.2, 0.3, and 0.5, respectively. We adopt intermediate classifiers with relatively low computational complexity compared to the backbone networks \cite{EEdesign}. Specifically, we assign more convolutional layers to the intermediate classifiers for a deeper backbone network, and the second early exit is lighter in computation than the first one. Since AlexNet is not deep, we insert two intermediate classifiers with 2 and 1 convolutional layer(s) respectively, followed by a max-pooling layer and an FC layer, which are placed after the first and second convolutional layer in the backbone network respectively. Besides, for VGG16-BN, the first intermediate classifier consists of 3 convolutional layers and 1 FC layer, which is inserted after the first convolutional layer in the backbone. The second intermediate classifier consists of 2 convolutional layers and 1 FC layer, and it is inserted after the third convolutional layer in the backbone. In addition, as ResNet44 applies the short connection structure, the early exits should avoid being inserted inside a residual block (RB). Hence, we insert the first intermediate classifier with 2 RBs, 1 max-pooling layer, and 1 FC layer after the first RB in the backbone, and the second early exit consists of 1 RB, 1 max-pooling layer, and 1 FC layer after the eighth RB in the backbone. TABLE \ref{table: two exit feature size} in APPENDIX A and TABLE \ref{table: two exit on device flops} in APPENDIX B respectively show the output feature size of each layer/RB in the early exits and the amount of computations required to process each part of the on-device network. We use stochastic gradient descent with a batch size of 128 and a weight decay of 0.0005 to train the early-exit networks. Cosine annealing is applied with an initial learning rate of 0.1 and an end learning rate of 0.0001 at the 200-th epoch. The total number of training epochs is 220. The hyper-parameter setting of training the Exit Predictor is similar to that of the early-exit networks except with a weight decay of 0.0002.

\begin{figure*}[htb]
  \centering
  \subfloat[AlexNet backbone on CIFAR10 with $\lambda_{1}=\lambda_{2}\in \{0.90, 0.92, 0.94, 0.97, 0.99\}$]{\includegraphics[width=0.3\textwidth]{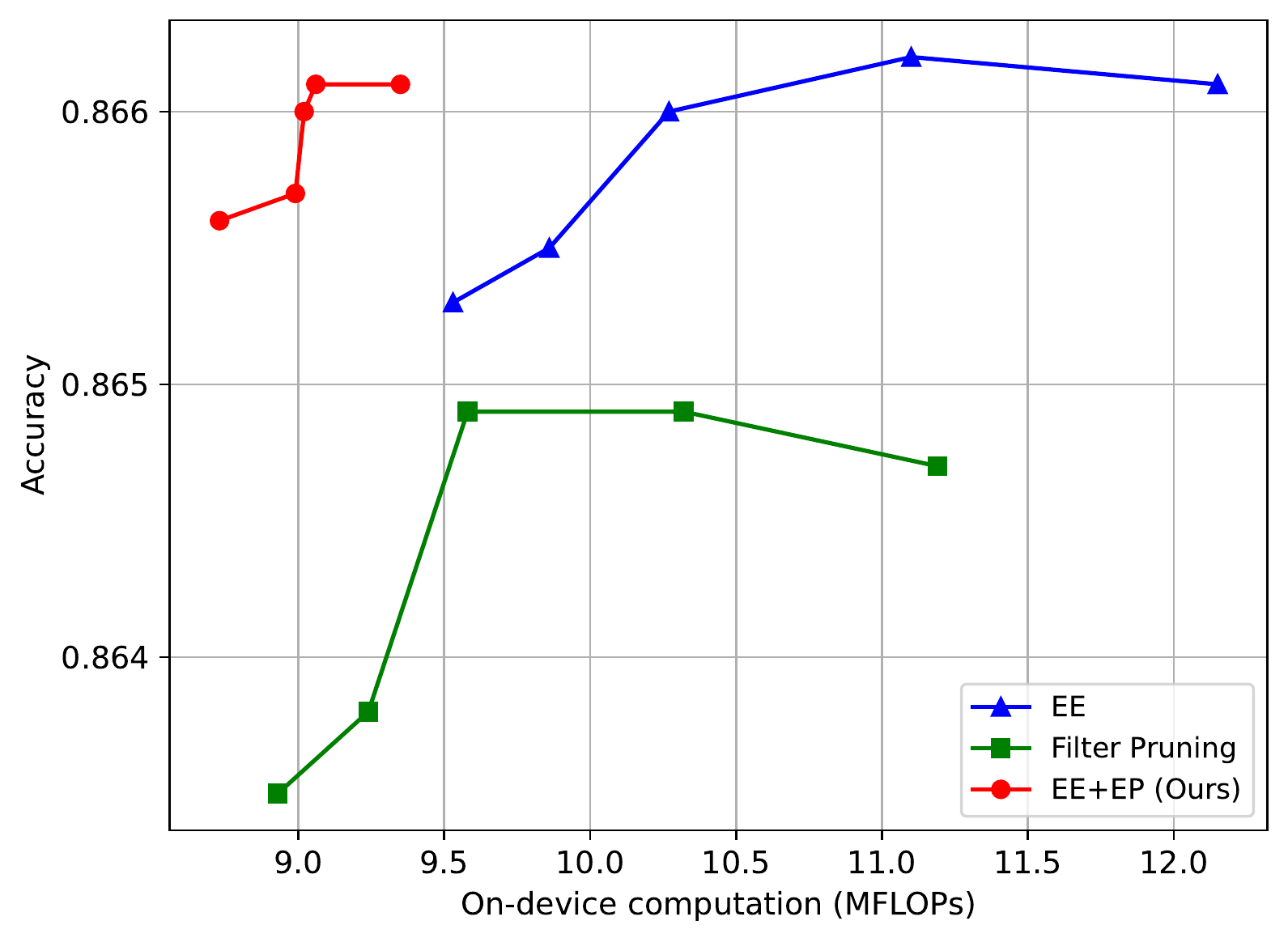}}
  \hfill
  \subfloat[VGG16-BN backbone on CIFAR10 with $\lambda_{1}=\lambda_{2}\in \{0.90, 0.95, 0.97, 0.99, 0.995\}$]{\includegraphics[width=0.3\textwidth]{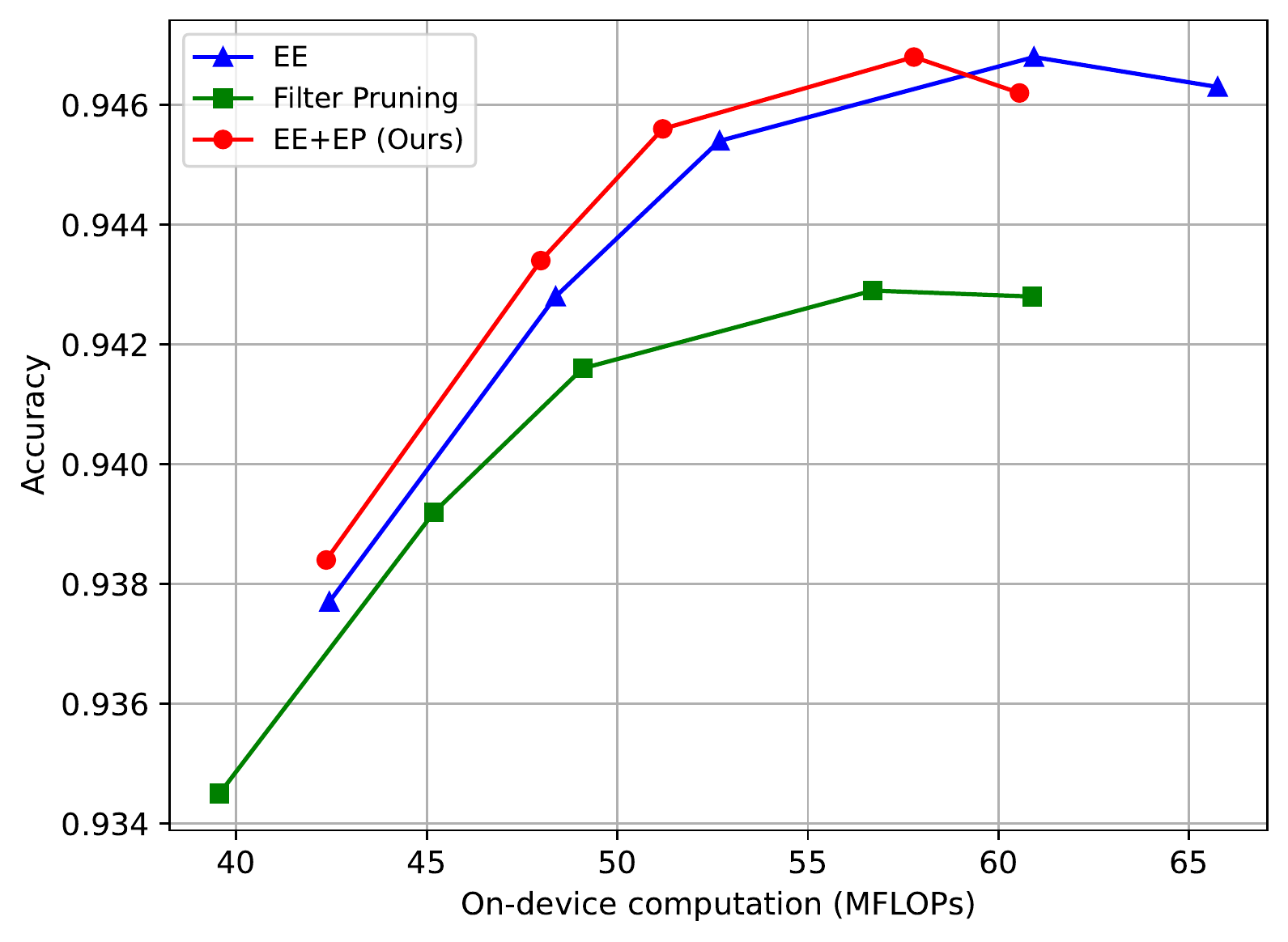}}
  \hfill
  \subfloat[ResNet44 backbone on CIFAR10 with $\lambda_{1}=\lambda_{2}\in \{0.85, 0.88, 0.93, 0.97, 0.99\}$]{\includegraphics[width=0.3\textwidth]{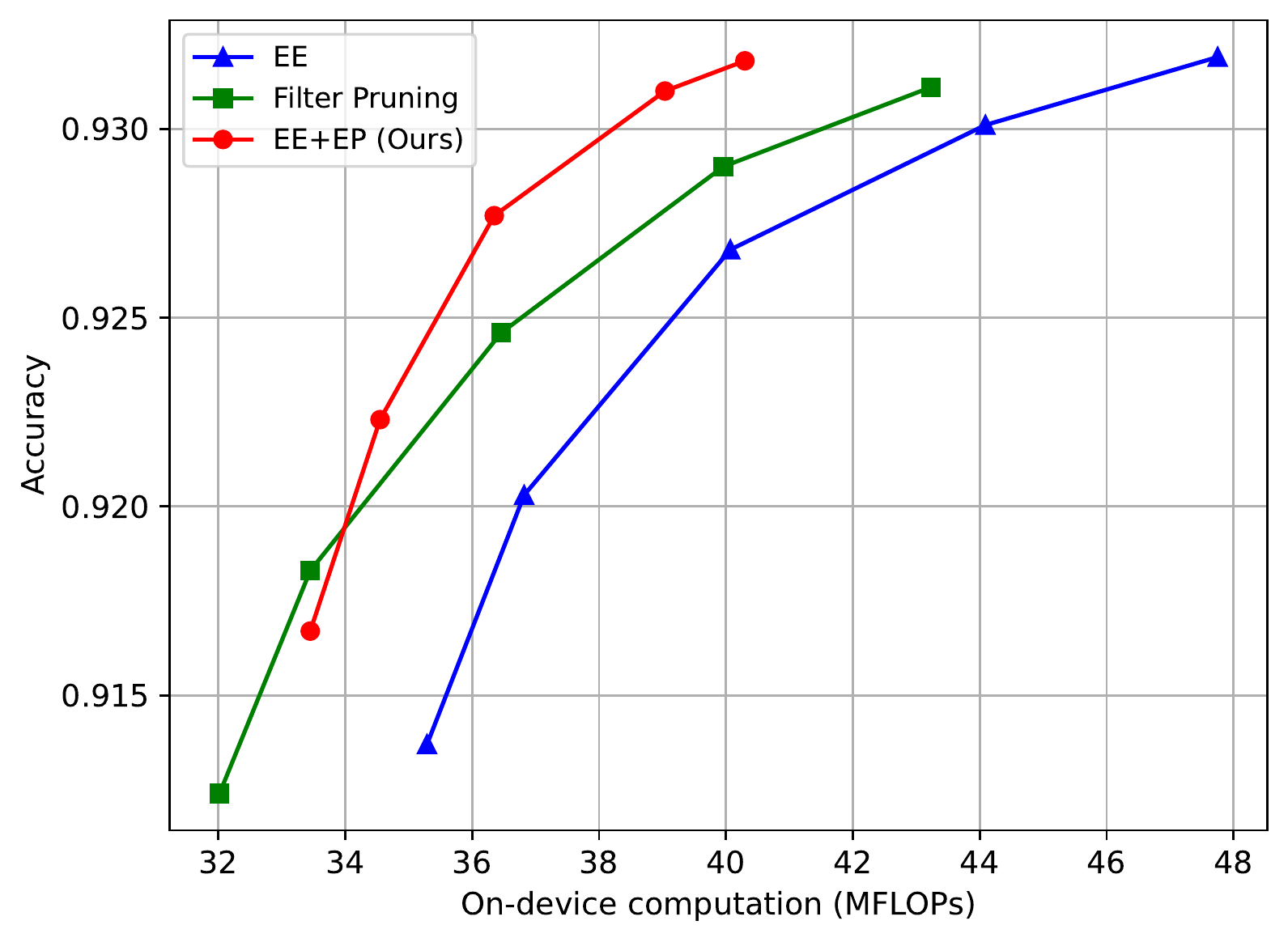}}\\
  \subfloat[AlexNet backbone on CIFAR100 with $\lambda_{1}=\lambda_{2}\in \{0.50, 0.55, 0.60, 0.70, 0.80\}$]{\includegraphics[width=0.3\textwidth]{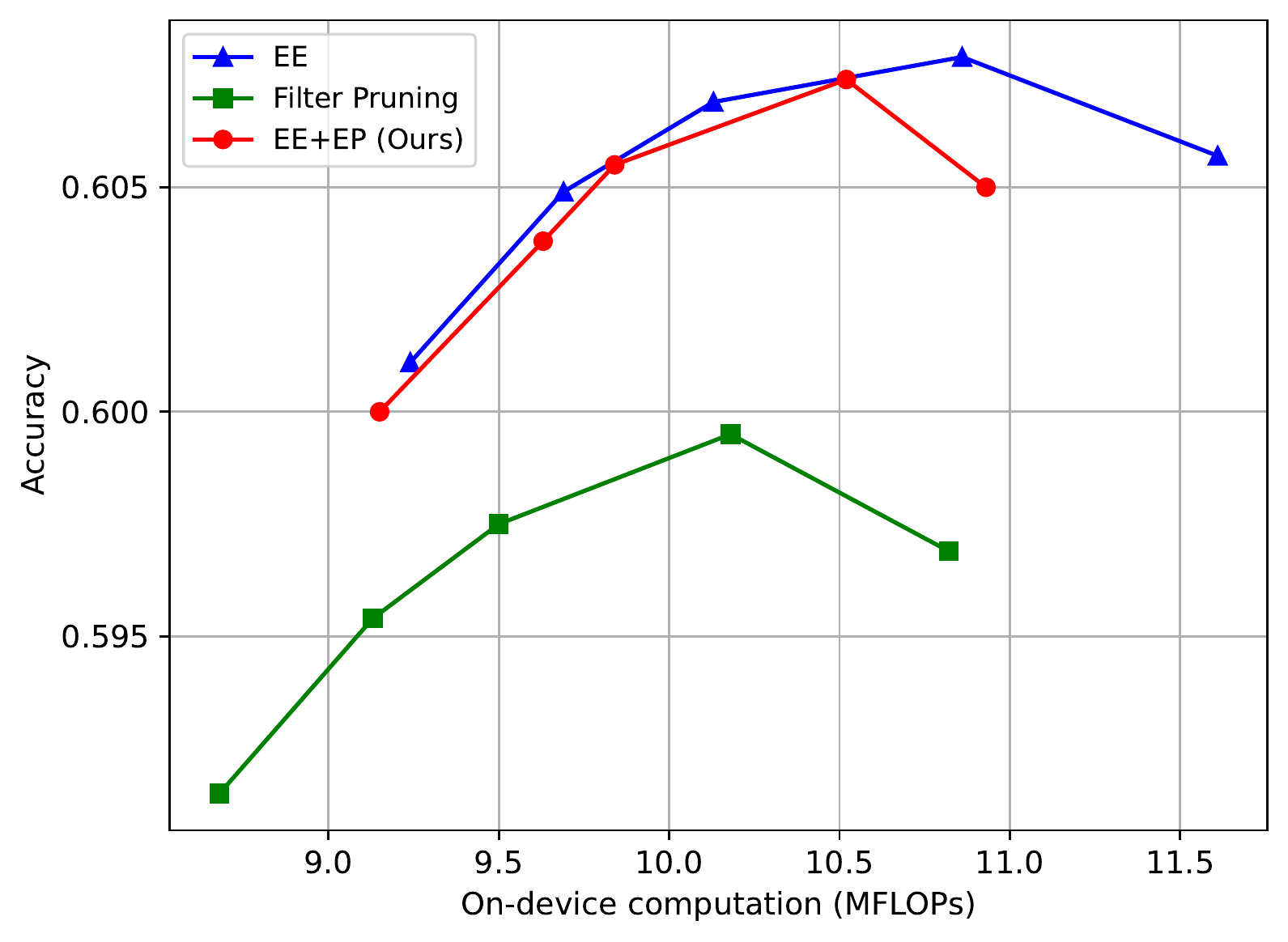}}
  \hfill
  \subfloat[VGG16-BN backbone on CIFAR100 with $\lambda_{1}=\lambda_{2}\in \{0.75, 0.80, 0.85, 0.90, 0.95\}$]{\includegraphics[width=0.3\textwidth]{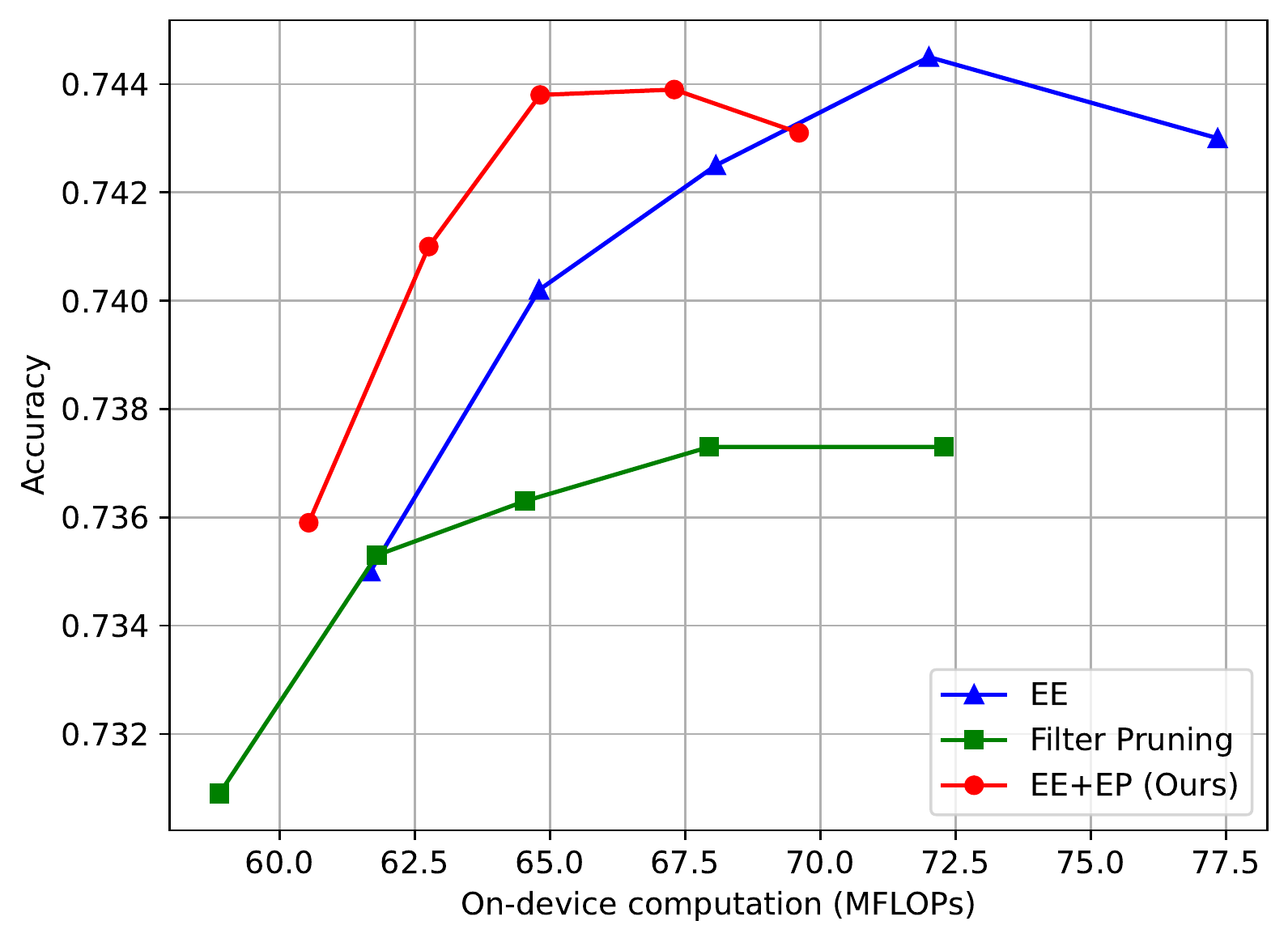}}
  \hfill
  \subfloat[ResNet44 backbone on CIFAR100 with $\lambda_{1}=\lambda_{2}\in \{0.70, 0.75, 0.80, 0.90, 0.95\}$]{\includegraphics[width=0.3\textwidth]{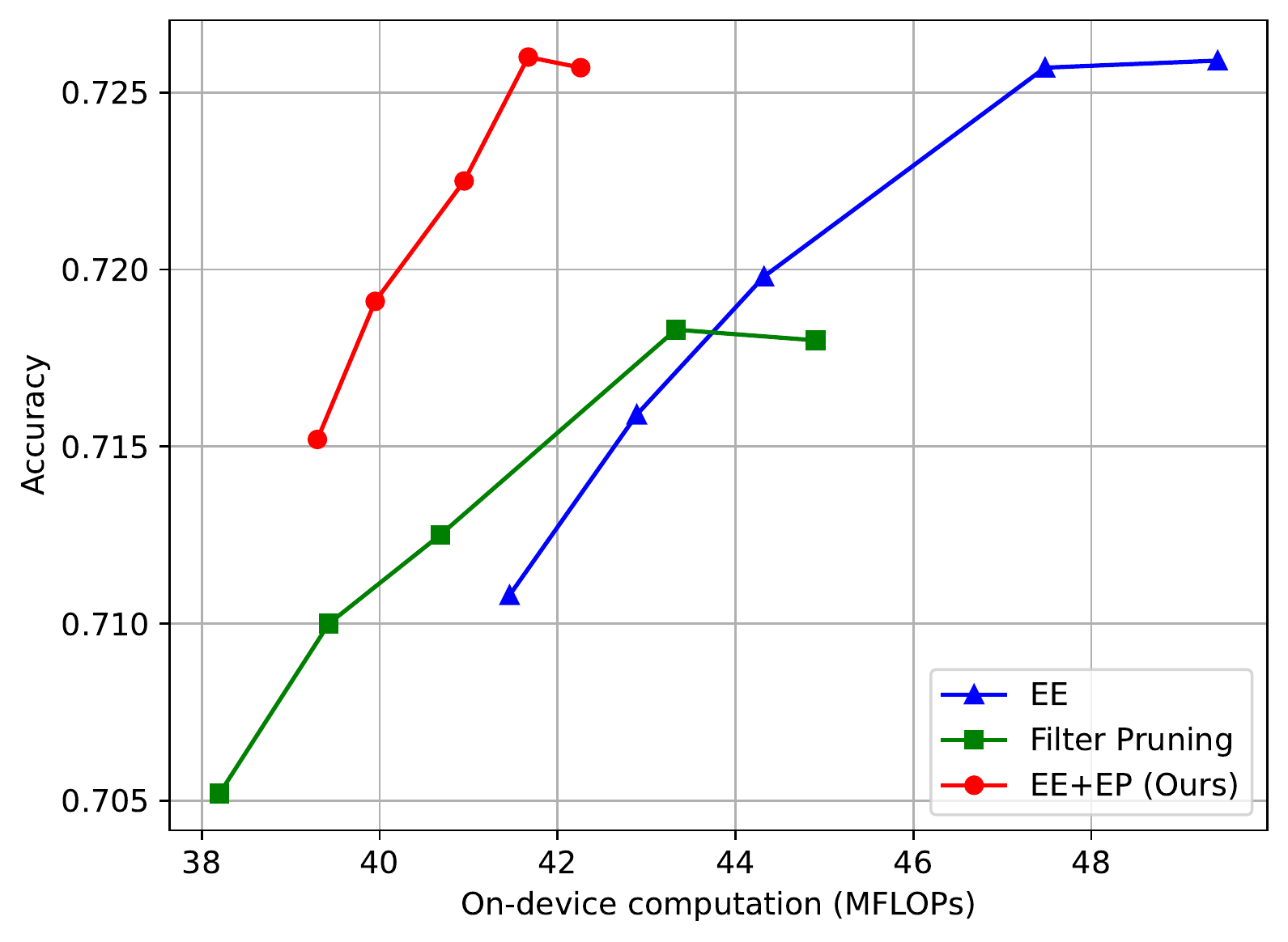}}
  \caption{Classification accuracy versus on-device computation overhead. The two confidence thresholds are assumed to be identical and increase with the on-device computation. The filter pruning method removes 10\% of the filters with the lowest $l_{2}$-norm values. The prediction thresholds $\bm{\gamma}$ are chosen as the ones with the lowest on-device computation while less than $2\%$ additional samples are terminated at the last exit, i.e., cannot be terminated by the two early exits.}
  \label{fig: predictive early exiting results}
\end{figure*}

\subsection{On-device Computation Savings}
\label{subsection: On-device Computation Savings}
We evaluate the performance of the Exit Predictor (EP) in terms of classification accuracy and average on-device computation for each inference data sample. For comparisons, the original early-exit (EE) network, as well as a filter pruning \cite{FilterPruning} method which removes the filters according to their $l_{2}$-norm values, are adopted as baselines. Since the edge server is well-resourced, filter pruning only applies to the on-device network (except the last FC layer to avoid drastic accuracy degradation).

\begin{figure}[ht]
    \centering
    \includegraphics[width=0.6\textwidth]{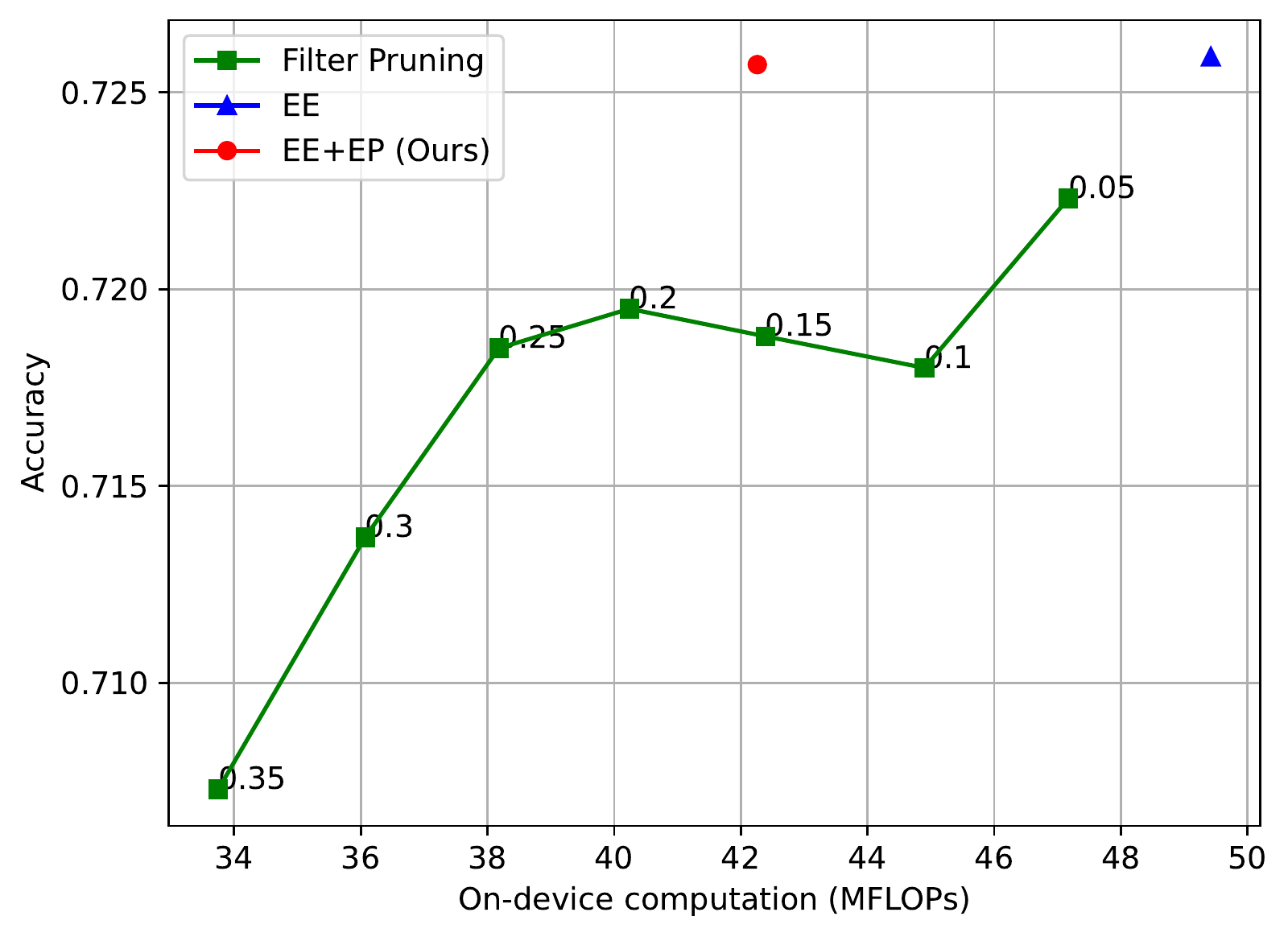}
    \caption{Classification accuracy versus on-device computation overhead for the early-exit network based on the ResNet44 backbone and the CIFAR100 dataset. The numbers in the figure indicate the pruning ratios of the filter pruning method. The two confidence thresholds of the early-exit network are both set as 0.95.}
    \label{fig: different prune ratio}
\end{figure}

In the experiments, we adjust the confidence thresholds of the early exits to achieve different tradeoff between accuracy and on-device computation overhead, and the results are shown in Fig. \ref{fig: predictive early exiting results}. In general, compared with the original early-exit networks, the proposed Exit Predictor can greatly reduce the on-device computation without much impairing the classification accuracy. Also, we observe that although the filter pruning method is able to save some on-device computations, it suffers from notable accuracy degradation. In other words, the proposed Exit Predictor is more suitable for scenarios with stringent accuracy requirements. As the values of the confidence thresholds become lower, the inference processes of more samples are terminated by the first early exit, leading to smaller on-device computation and reduced inference accuracy. Thus, fewer samples are able to skip the first early exit so that the on-device computation saving brought by the Exit Predictor becomes less significant. There is an interesting observation from the results of the early-exit networks with the AlexNet and VGG16-BN backbones that the accuracy drops even with high on-device computation overhead, i.e., large values of the confidence thresholds, which can be explained by the overthinking of early-exit networks \cite{EEnetOverthink}. Fig. \ref{fig: different prune ratio} further shows the performance of the filter pruning method with different pruning ratios. We find that the proposed early exit prediction mechanism still substantially outperforms the filter pruning method.

\begin{figure}[htb]
    \centering
    \includegraphics[width=0.6\textwidth]{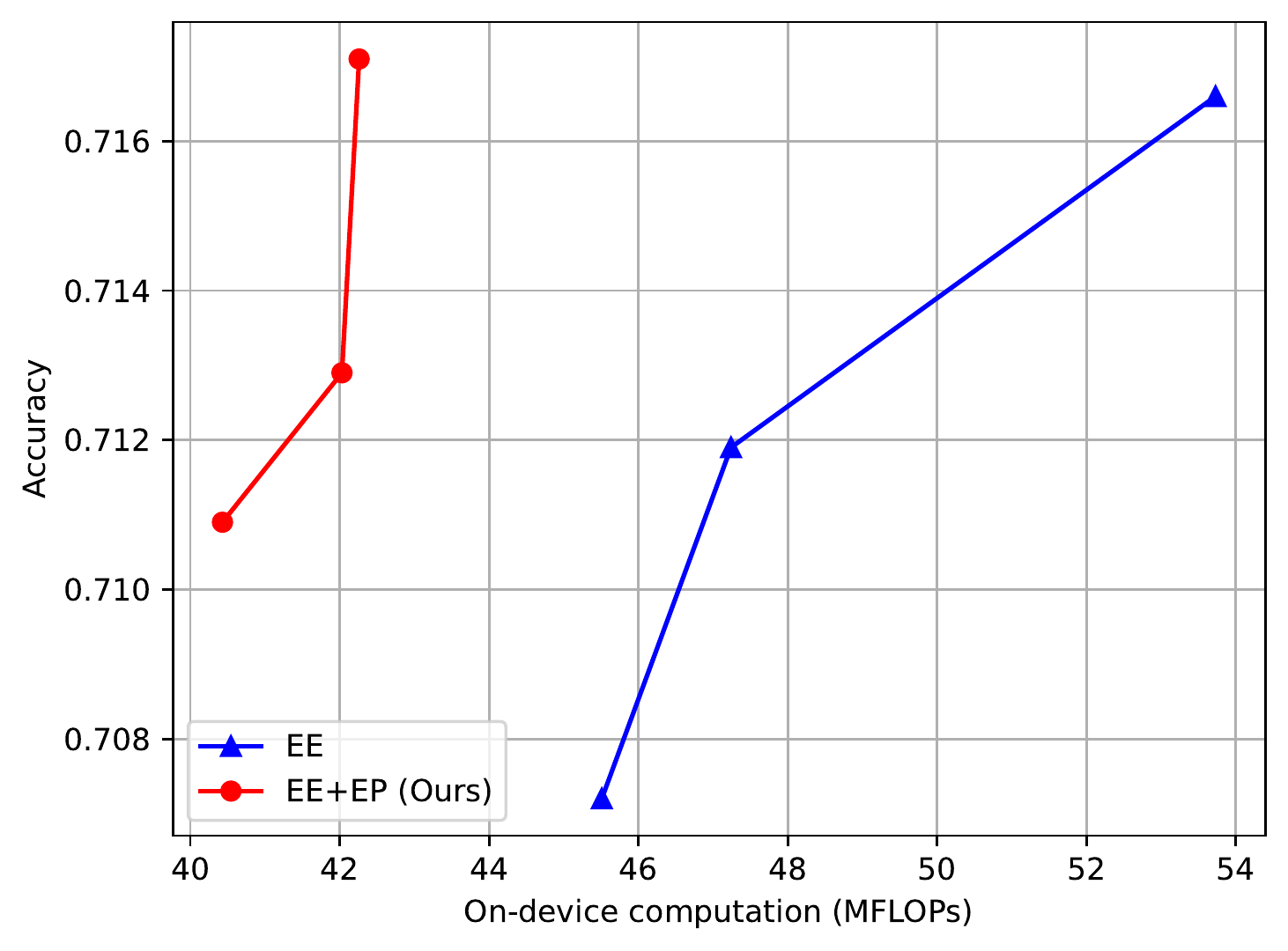}
    \caption{Classification accuracy versus on-device computation overhead on the CIFAR100 dataset. The early-exit network is constructed by inserting three early exits to the ResNet44 backbone, with weights in (\ref{equation: ee network loss function}) for the four exits as 0.2, 0.2, 0.2, 0.4. In this figure, the confidence thresholds are set as $\lambda_{1}=\lambda_{2}=\lambda_{3}=\{0.75, 0.8, 0.95\}$.}
    \label{fig: ResNet44 with 3 exits}
\end{figure}

\textbf{\textit{Ablation Studies}}: We first examine the performance of different Exit Predictor architectures on the CIFAR100 dataset, as shown in TABLE \ref{table: predictor type}, including the proposed design in Fig. 6 (``Ours"), the proposed design in Fig. 6 without the channel concatenation and the squeeze-and-excitation operations (``w/o Cat. \& SE"), the network architecture consists of only two FC layers (``FCs only"), and the LeNet \cite{Lenet}. From this set of results, we find all the considered network architectures achieve similar classification accuracy, while the proposed network architecture for the Exit Predictor always results in the lowest on-device computation overhead. We further test the Exit Predictor for early-exit networks with three early exits. We choose the ResNet44 as the backbone, which is deeper than AlexNet and VGG16-BN so that more early exits can be inserted. TABLE \ref{table: three exit feature size} in APPENDIX A and TABLE \ref{table: three exit on device flops} in APPENDIX B respectively show the output feature size of each residual block in the early exits and the amount of computations required to process each part of the on-device network. The classification accuracy versus the on-device computation overhead is shown in Fig. \ref{fig: ResNet44 with 3 exits}, from which, we can still observe significant improvements achieved by the Exit Predictor compared with the original early-exit networks.

\begin{table*}[ht]
\caption{Performance of Exit Predictors with different network architectures on the CIFAR100 dataset.}
\label{table: predictor type}
\centering
\tiny
\begin{tabular}{cccc|cccc|cccc}
\hline
\multicolumn{4}{c|}{Backbone: AlexNet}  & \multicolumn{4}{c|}{Backbone: VGG16-BN} & \multicolumn{4}{c}{Backbone: ResNet44}  \\ \hline
\multicolumn{1}{c}{\begin{tabular}[c]{@{}c@{}}EP\\ Architecture\end{tabular}} & \multicolumn{1}{c}{Acc.(\%)} & \multicolumn{1}{c}{\begin{tabular}[c]{@{}c@{}}On-device\\ FLOPs\end{tabular}} & \begin{tabular}[c]{@{}c@{}}EP\\ FLOPs\end{tabular} & \multicolumn{1}{c}{\begin{tabular}[c]{@{}c@{}}EP\\ Architecture\end{tabular}} & \multicolumn{1}{c}{Acc.(\%)}            & \multicolumn{1}{c}{\begin{tabular}[c]{@{}c@{}}On-device\\ FLOPs\end{tabular}} & \begin{tabular}[c]{@{}c@{}}EP\\ FLOPs\end{tabular} & \multicolumn{1}{c}{\begin{tabular}[c]{@{}c@{}}EP\\ Architecture\end{tabular}} & \multicolumn{1}{c}{Acc.(\%)}            & \multicolumn{1}{c}{\begin{tabular}[c]{@{}c@{}}On-device\\ FLOPs\end{tabular}} & \begin{tabular}[c]{@{}c@{}}EP\\ FLOPs\end{tabular} \\ \hline
\multicolumn{1}{c}{Ours}                                                          & \multicolumn{1}{c}{60.74} & \multicolumn{1}{c}{\textbf{10.52}} & 0.40  & \multicolumn{1}{c}{Ours}  & \multicolumn{1}{c}{74.39} & \multicolumn{1}{c}{\textbf{67.30}}                                            & 0.40                                                           & \multicolumn{1}{c}{Ours}                                                          & \multicolumn{1}{c}{72.60} & \multicolumn{1}{c}{\textbf{41.67}}                                            & 0.40                                                           \\ \hline
\multicolumn{1}{c}{w/o Cat. \& SE}                                                & \multicolumn{1}{c}{60.66}          & \multicolumn{1}{c}{10.59}                                                     & \textbf{0.29}                                                  & \multicolumn{1}{c}{w/o Cat. \& SE}                                                & \multicolumn{1}{c}{74.49}          & \multicolumn{1}{c}{67.65}                                                     & \textbf{0.29}                                                  & \multicolumn{1}{c}{w/o Cat. \& SE}                                                & \multicolumn{1}{c}{72.57}          & \multicolumn{1}{c}{41.71}                                                     & \textbf{0.29}                                                  \\ \hline
\multicolumn{1}{c}{FCs only}  & \multicolumn{1}{c}{\textbf{60.77}}  & \multicolumn{1}{c}{11.24} & 0.39  & 
\multicolumn{1}{c}{FCs only}                                                      & \multicolumn{1}{c}{\textbf{74.51}}          & \multicolumn{1}{c}{70.38}                                                     & 0.39                                                           & \multicolumn{1}{c}{FCs only}                                                      & \multicolumn{1}{c}{\textbf{72.62}}          & \multicolumn{1}{c}{42.71}                                                     & 0.39                                                           \\ \hline
\multicolumn{1}{c}{LeNet}                                                         & \multicolumn{1}{c}{60.74}          & \multicolumn{1}{c}{11.17}                                                     & 0.67                                                           & \multicolumn{1}{c}{LeNet}                                                         & \multicolumn{1}{c}{74.46}          & \multicolumn{1}{c}{68.46}                                                     & 0.67                                                           & \multicolumn{1}{c}{LeNet}                                                         & \multicolumn{1}{c}{72.61}          & \multicolumn{1}{c}{42.21}                                                     & 0.67                                                           \\ \hline
\end{tabular}
\end{table*}

\subsection{Latency and Accuracy under Different Communication Bandwidths}
\label{subsection: Latency and Accuracy under Different Communication Bandwidths}
The experiments in this section are based on the VGG16-BN backbone and the CIFAR100 dataset. The architecture of the feature encoder and decoder are adopted from \cite{bottlenet++} by substituting the Sigmoid function in the encoder to a ReLU function, which reduces the height and width of each channel by half, and the number of channels to $\frac{1}{4}$. Besides, the quantization operation converts the bit-width from 32 to 8. Therefore, the feature compression ratio is given by 64, resulting in less than 0.5\% accuracy degradation. We choose the Raspberry Pi 3 as the edge device, which offers a computation speed of 3.62 GFLOPS\footnote{\href{https://web.eece.maine.edu/~vweaver/group/green_machines.html}{https://web.eece.maine.edu/$\sim$vweaver/group/green\_machines.html}}. The latency requirement is set to 30 ms. We train 3 regression models respectively for the bandwidth intervals 0.1-1 Mbit/s, 1-10 Mbit/s, and 1-100 Mbit/s in this experiment. The sets of communication bandwidth values to train the three regression models are given by \{0.1, 0.3, 0.5, 0.7, 1\}, \{1, 3, 5, 7, 10\} and \{10, 30, 50, 70, 100\} Mbit/s respectively. We compare the proposed method (``EE+EP (Adaptive)") in Section \ref{section latency-aware early exit prediction} with the following baselines (Feature compression is applied to all the methods for fair comparisons):
\begin{itemize}
    \item Backbone: This method uses the VGG16-BN backbone for device-edge co-inference.
    \item EE (Static): This method refers to the original early-exit network, where the confidence scores of the two early exits are set as 0.95 and 0.85 to achieve the best accuracy.
    \item EE+EP (Static): This method applies the Exit Predictor proposed in Section \ref{section the proposed early exit prediction mechanism} to the early-exit network in the EE (Static) method, where $\{\gamma_{i}\}$'s are chosen with the same criterion as in Fig. \ref{fig: predictive early exiting results}.
    \item EE (Adaptive): This method adapts the confidence thresholds of an early-exit network according to the communication bandwidth to meet the latency requirement.
    \item EE (Adaptive, Pruning): On top of the EE (Static) method, this method further applies filter pruning to the convolutional layer in the feature encoder to reduce the number of channels to be transmitted. The pruning ratio is chosen such that the latency requirement is not violated.
    \item EE+Oracle (Adaptive): This is an idealized method and assumes that there is an oracle that can perfectly guide the samples to be computed by the exit that can obtain a confident inference result, i.e., $c_{n}\geq \lambda_{n}$. In other words, it can be regarded as a performance upper bound of the proposed EE+EP (Adaptive) method.
\end{itemize}

\begin{figure}[htb]
    \centering
    \subfloat[Latency versus communication bandwidth]{\includegraphics[width=0.6\textwidth]{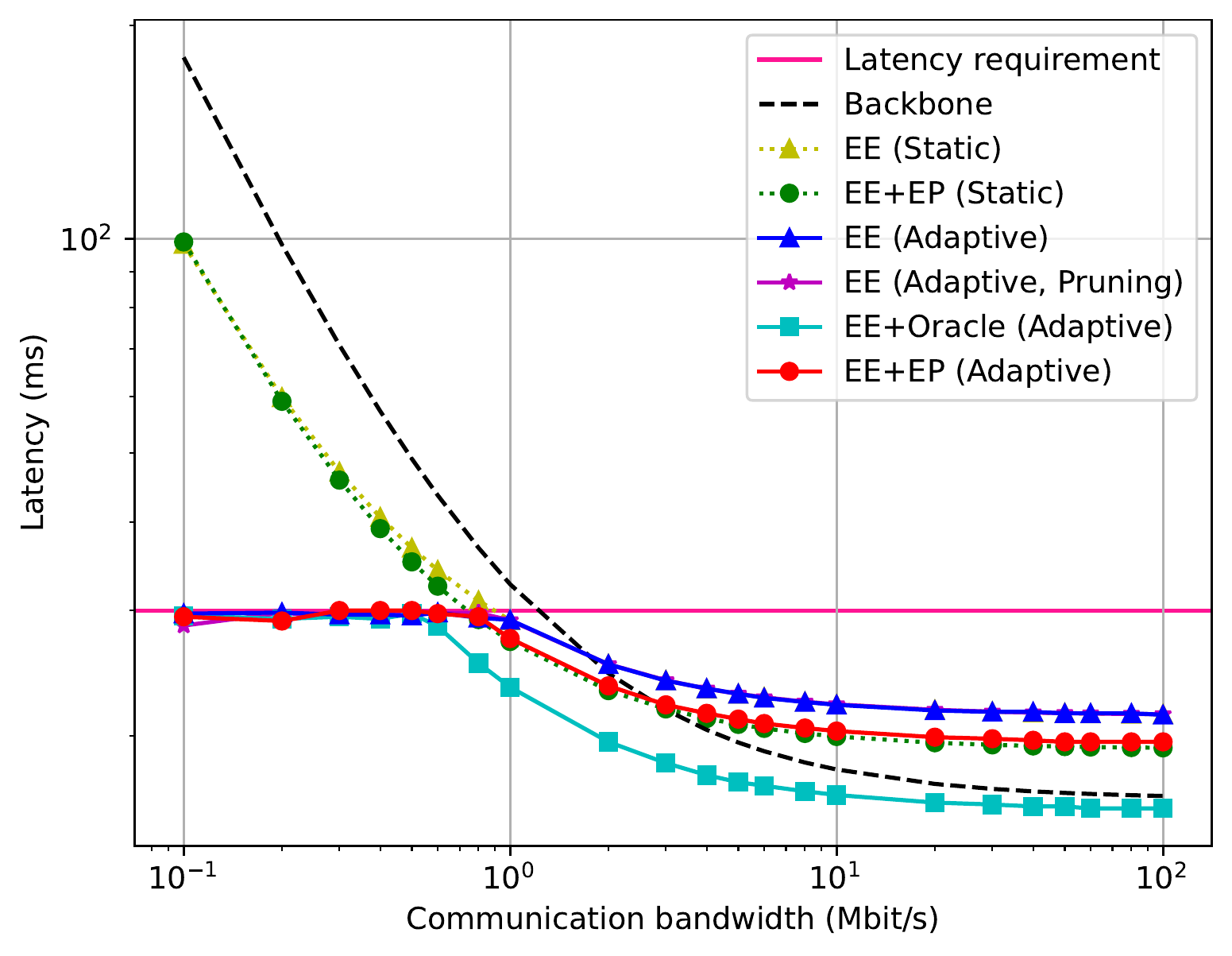}}
    \\
    \subfloat[Classification accuracy versus communication bandwidth]{\includegraphics[width=0.6\textwidth]{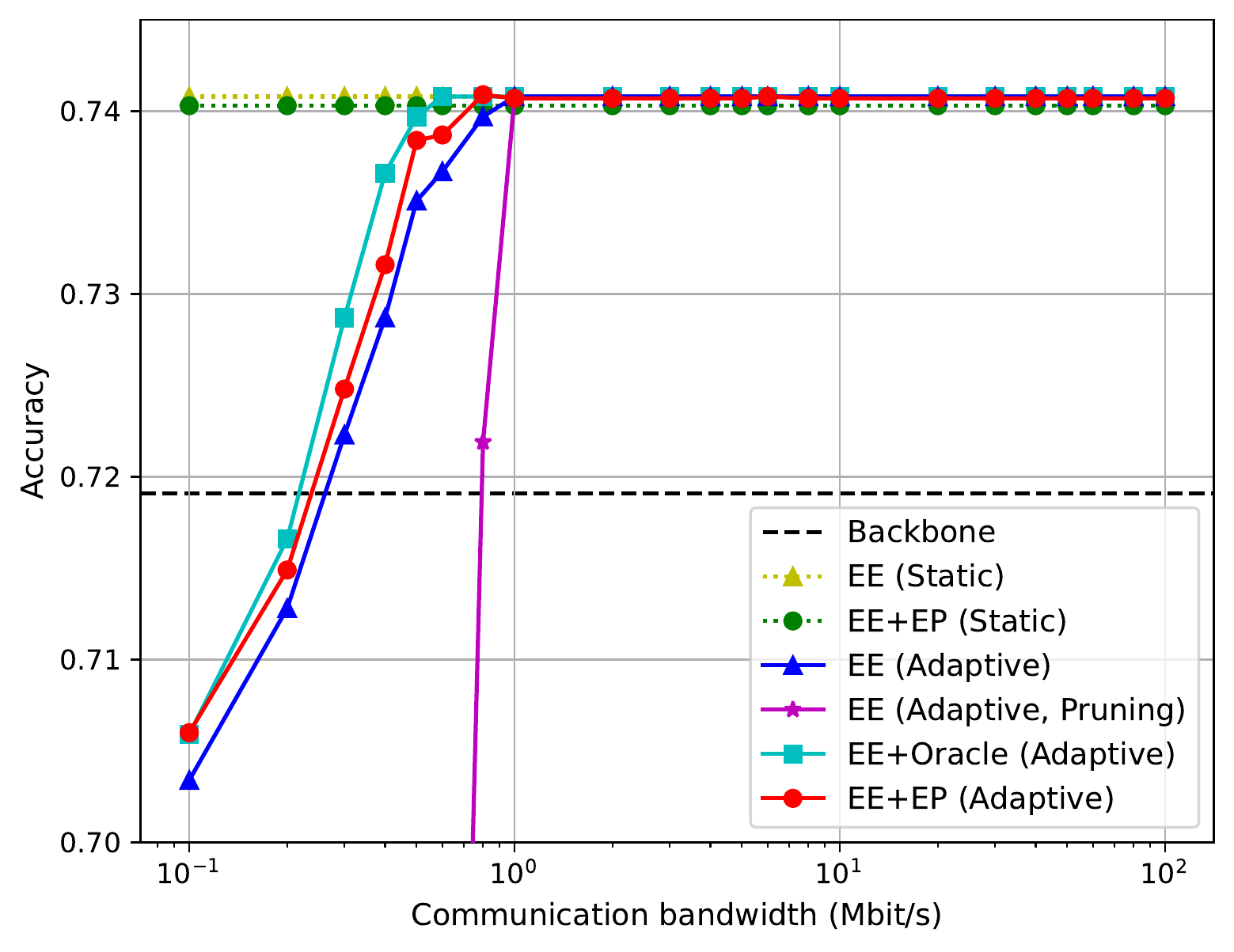}}
    \caption{Latency and classification accuracy under different communication bandwidths with the VGG16-BN backbone on the CIFAR100 dataset.}
    \label{fig: adaptive}
\end{figure}

The latency and accuracy performance under different communication bandwidths are shown in Fig. \ref{fig: adaptive}. It can be observed from the figure that, when the bandwidth is low, the latency requirement can not be satisfied by the three static methods, despite they are able to achieve the best classification accuracy. Compared with the early-exit network, the end-to-end inference latency of the VGG16-BN backbone is significant higher when the communication bandwidth is low because all the samples need to be processed by the server-based network. When the communication bandwidth is high, the VGG16-BN backbone achieves lower latency compared with the early-exit network as the computation latency dominates in this regime. This result demonstrates the benefits of early-exit networks for device-edge co-inference systems with insufficient communication resources. Kindly note that the VGG16-BN backbone has slight accuracy degradation compared with the early exit network\footnote{Similar observations can be found in \cite{branchynet, dynexit}.}. In contrast, the four adaptive methods can always meet the inference latency requirement. However, the EE (Adaptive, Pruning) method suffers from serious accuracy degradation as too many filters are removed from the encoded feature in order to meet the latency requirement. In the low-bandwidth regime (i.e., 0.1 to 1 Mbit/s), the proposed EE+EP (Adaptive) method outperforms the EE (Adaptive) baseline, which is because on-device computation latency reduction of the ``easy'' samples brought by the EP so that more ``hard'' samples can be inferred by the last exit with higher accuracy. The idealized EE+Oracle (Adaptive) method achieves both latency and accuracy improvements over our proposed method because it can precisely avoid unnecessary computation. When the communication bandwidth is higher than 1 Mbit/s, all the adaptive methods converge to the highest accuracy, and with the early exit prediction mechanism, lower latency can be achieved for its strength in saving the on-device computation.
\section{Conclusions}
\label{section conclusions}
In this paper, we considered a device-edge co-inference system empowered by early-exit networks and proposed a low-complexity early exit prediction mechanism, named Exit Predictor, to improve the on-device computation efficiency. Assisted by learning-based feature compression, we extended the early-exit network for latency-aware edge inference under different bandwidth conditions. Experiment results demonstrated the benefits of our proposed methods in reducing the on-device computation overhead and end-to-end inference latency for resource-constrained edge AI.

However, as can be seen from the experiment results, the proposed method for latency-aware edge inference still has a considerable performance gap compared to the idealized method, implying there is room for further improvement. Besides, there are many future directions beyond this study. For example, it would be interesting to combine model compression and layer-skipping techniques with the proposed early exit prediction mechanism, and extend our investigation to multi-device cooperative edge inference.

{\appendices
\section*{Appendix A}
\label{section appendix a}
\section*{Output Feature Size of the Early Exit Layers/Residual Blocks}
\begin{table}[H]
\footnotesize
\centering
\caption{Output feature size (channel$\times$height$\times$width) of each layer/residual block in the early exits (for early-exit networks with two intermediate classifiers).}
\label{table: two exit feature size}
\begin{tabular}{c|c|c}
\hline
Backbone & Exit 1                                                                                                                     & Exit 2                                                                                    \\ \hline
AlexNet  & \begin{tabular}[c]{@{}c@{}}$64\times 8\times 8$\\      $64\times 8\times 8$\end{tabular}                                   & $64\times 4\times 4$                                                                      \\ \hline
VGG16-BN & \begin{tabular}[c]{@{}c@{}}$64\times 16\times 16$\\      $32\times 16\times 16$\\      $32\times 16\times 16$\end{tabular} & \begin{tabular}[c]{@{}c@{}}$128\times 8\times 8$\\      $64\times 8\times 8$\end{tabular} \\ \hline
ResNet44 & \begin{tabular}[c]{@{}c@{}}$16\times 32\times 32$\\      $16\times 32\times 32$\end{tabular}                               & $32\times 16\times 16$                                                                    \\ \hline
\end{tabular}
\end{table}

\begin{table}[H]
\footnotesize
\centering
\caption{Output feature size (channel$\times$height$\times$width) of each residual block in the early exits (for the early-exit network with three intermediate classifiers in the ablation study).}
\label{table: three exit feature size}
\begin{tabular}{c|c|c|c}
\hline
Backbone & Exit 1  & Exit 2 & Exit 3  \\ \hline
ResNet44 & \begin{tabular}[c]{@{}c@{}}$16\times 32\times 32$\\      $16\times 32\times 32$\end{tabular} & $16\times 32\times 32$ & $32\times 16\times 16$ \\ \hline
\end{tabular}
\end{table}

\section*{Appendix B}
\label{section appendix b}
\section*{Amount of Computations Required to Process Each Part of the On-device Network}

\begin{table}[H]
\footnotesize
\centering
\caption{Amount of computations (in MFLOPs) required for each part of the on-device network (for early-exit networks with two intermediate classifiers).}
\label{table: two exit on device flops}
\begin{tabular}{c|c|c|c|c|c}
\hline
Dataset                   & Backbone & $O_{l_{1}}$ & $O_{e_{1}}$ & $O_{l_{2}}$ & $O_{e_{2}}$ \\ \hline
\multirow{3}{*}{CIFAR10}  & AlexNet  & 0.49                 & 4.75                 & 7.11                 & 1.78                 \\ \cline{2-6} 
                          & VGG16-BN & 1.97                 & 16.70                & 56.98                & 14.23                \\ \cline{2-6} 
                          & ResNet44 & 5.29                 & 9.66                 & 32.53                & 4.79                 \\ \hline
\multirow{3}{*}{CIFAR100} & AlexNet  & 0.49                 & 4.84                 & 7.11                 & 1.80                 \\ \cline{2-6} 
                          & VGG16-BN & 1.97                 & 17.43                & 56.98                & 14.60                \\ \cline{2-6} 
                          & ResNet44 & 5.29                & 10.03                 & 32.53                & 4.97                 \\ \hline
\end{tabular}
\end{table}

\begin{table}[H]
\footnotesize
\centering
\caption{Amount of computations (in MFLOPs) required for each part of the on-device network (for the early-exit network with three intermediate classifiers in the ablation study).}
\label{table: three exit on device flops}
\begin{tabular}{cccccc}
\hline
\multicolumn{6}{c}{Backbone: ResNet44, Dataset: CIFAR100}  \\ \hline
\multicolumn{1}{c|}{$O_{l_{1}}$} & \multicolumn{1}{c|}{$O_{e_{1}}$} & \multicolumn{1}{c|}{$O_{l_{2}}$} & \multicolumn{1}{c|}{$O_{e_{2}}$} & \multicolumn{1}{c|}{$O_{l_{3}}$} & $O_{e_{3}}$ \\ \hline
\multicolumn{1}{c|}{5.29} & \multicolumn{1}{c|}{10.03} & \multicolumn{1}{c|}{14.4} & \multicolumn{1}{c|}{5.23} & \multicolumn{1}{c|}{18.13} & 4.97 \\ \hline
\end{tabular}
\end{table}}

\printbibliography

\vfill

\end{document}